%% file: neurips.tex
\documentclass{article}
\usepackage{arydshln}
\usepackage{float}  
\usepackage{subfigure} 
\usepackage{verbatim}
\usepackage[toc,page]{appendix}
\usepackage[table,dvipsnames]{xcolor}
\definecolor{ref-blue}{rgb}{0.0, 0.24, 0.95}
\definecolor{c1st}{HTML}{79AC78}

\usepackage[
    colorlinks=true,
    citecolor={ref-blue},
    urlcolor={ref-blue}
]{hyperref}

\newcommand*{\preprint}{}%

\ifdefined\preprint
    \usepackage[preprint,nonatbib]{neurips}
\else
    \usepackage{neurips}
\fi

\usepackage{amsmath,amssymb}

\usepackage{multirow, booktabs, makecell}

\usepackage{graphics,overpic}

\usepackage{listings}

\usepackage[utf8]{inputenc} 
\usepackage[T1]{fontenc}    
\usepackage{hyperref}       
\usepackage{url}            
\usepackage{amsfonts}       
\usepackage{nicefrac}       
\usepackage{microtype}      

\newcommand{\mypar}[1]{\textbf{#1}\hspace{0.5em}}
\newcommand{\best}[1]{\cellcolor{c1st!30}#1}
\newcommand{\base}[1]{{\color{gray}{#1}}}
\newcommand{\zuojie}[1]{{\color{blue}{#1}}}

\usepackage{cleveref}
\crefname{equation}{Eq.}{Eq.}
\crefname{figure}{Fig.}{Fig.}
\crefname{table}{Tab.}{Tab.}
\crefname{section}{Sec.}{Sec.}

\title{
    Dominant Shuffle:
    A Simple Yet Powerful Data Augmentation for
    Time-series Prediction
}

\author{%
  Kai Zhao$^{1}$ \quad Zuojie He$^{2}$ \quad Alex Hung$^{1}$ \quad Dan Zeng$^{2}$ \\
  $^1$UCLA \quad $^2$ Shanghai University \\
  \texttt{kz@kaizhao.net}
}

\begin{document}

\maketitle

\begin{abstract}
    \input{abstract1.tex}
    \ifdefined\preprint
    Code can be accessed at \url{https://kaizhao.net/time-series}.
    \fi
\end{abstract}

\section{Introduction}\label{sec:intro}
\input{introduction/nips.tex}

\section{Related Work}
\input{related-work/nips.tex}

\section{Dominant Frequency Shuffle for Time-series}
\input{method/nips.tex}

\section{Experiments}
In this section,
we first introduce the implementation details in~\cref{sec:impl-details},
and then compared the performance of various SOTA
models with and without dominant shuffle in~\cref{sec:sota}.
In~\cref{sec:exp-compare-with-other-da}, we thoroughly compared dominant shuffle with various data augmentation methods.
Finally, we conducted ablation studies to verify hyperparameter sensitivity and justify design choices in~\cref{sec:ablation}.

\subsection{Experimental Setups}\label{sec:impl-details}
\mypar{Implementation details}
All the experiments were conducted with the PyTorch~\cite{paszke2019pytorch} framework
on a single NVIDIA RTX 3090 GPU. 
Some of the experimental results were
from respective original papers,
and some were reproduced
using official code with default configurations.
We only changed the data augmentation for fair comparisons.
Please refer to~\cref{appd-impl-details}
for the details about our reimplementations.
%
Following the practice of~\cite{chen2023fraug},
we performed data augmentations to double the size of the original training dataset unless otherwise specified.


\mypar{Evaluation protocols}
We tested our method with
short-term and long-term prediction protocols.
In the long-term protocol, the prediction period $T$ ranges from 96 to 720, with variations at 96, 192, 336, and 720. 
In contrast, 
the short-term protocol has prediction periods ranging from 12 to 48, with variations at 12, 24, 36, and 48.
Following the common practice of previous works
~\cite{zhou2021informer,wu2021autoformer,zhou2022fedformer,liu2024itransformer,wang2023micn,wu2023timesnet},
we quantified the performance of the prediction
using the mean-squared error (MSE) between the ground-truth
and the prediction.

\mypar{Datasets}
For long-term prediction, 
we experimented on eight well-established benchmarks:
the ETT datasets (ETTh1, ETTh2, ETTm1, ETTm2)~\cite{zhou2021informer},
and the
Weather, Electricity, Exchange, and Traffic datasets~\cite{wu2021autoformer}.
For short-term prediction,
following iTransformer~\cite{liu2024itransformer},
we used four 
public traffic network datasets (PEMS03, PEMS04, PEMS07, PEMS08)
from PEMS~\cite{chen2001freeway}.

Each dataset is divided into training, testing, and evaluation subsets in specific ratios.
The training, testing, and evaluation ratio is 6:2:2
for ETT and PEMS datasets,
and the ratio is 7:1:2 for Electricity, Traffic, Weather, and Exchange-rate datasets.
Detailed statistics of these datasets are summarized in~\cref{appd:details-datasets}.
For each setting (dataset+prediction length $T$),
we tuned the optimal number of dominant frequencies $k$
on the evaluation set.
The optimal $k$ on various datasets can be found in
~\cref{appd:sec-opt-k}.

\mypar{Baseline Models}
We selected diverse
models as the baseline in our experiments,
including two Transformer-based
(iTransformer~\cite{liu2024itransformer},
Autoformer~\cite{wu2021autoformer}),
two MLP-based methods 
(TiDE~\cite{das2023longterm}, Lightts~\cite{zhang2022less}),
and two temporal convolutional network (TCN) based methods
(MICN~\cite{wang2023micn}, SCINet~\cite{liu2022scinet}).
iTransformer (Liu et al., 2024) is the state-of-the-art in Transformer-based models,
TiDE (Das et al., 2023) is the state-of-the-art MLP-based model, 
and MICN (Wang et al., 2023) is the state-of-the-art TCN-based model.
For short-term prediction,
we used the SOTA iTransformer~\cite{liu2024itransformer} model
on PEMS~\cite{chen2001freeway} dataset as the baseline model.


\mypar{Other data augmentation methods}
We compared the proposed method
with nine existing data augmentation methods,
including three time-domain augmentations
(ASD~\cite{forestier2017generating}, MSB~\cite{bandara2021improving}
Upsample ~\cite{semenoglou2023data}),
five frequency-domain methods
(FreqMix~\cite{chen2023fraug}, 
FreqMask~\cite{chen2023fraug}, 
FreqAdd~\cite{zhang2022self}, 
FreqPool~\cite{chen2023supervised},
Robusttad~\cite{gao2020robusttad}),
and a temporal-frequency method STAug~\cite{zhang2023towards}.

\subsection{Comparison With State-of-the-arts}\label{sec:sota}
We first compared our method with other
state-of-the-art time series prediction models published in top-tier venues.
We compared the performance of
recent models
(iTransformer~\cite{liu2024itransformer} (ICLR2024),
SCINet~\cite{liu2022scinet} (NIPS2022)
AutoFormer~\cite{wu2021autoformer} (NIPS2021))
with and without dominant shuffle.
The averaged mean squared errors (MSE) across various
prediction lengths (96, 192, 336, 720)
is calculated for each dataset.
\begin{figure}[!htb]
     \begin{overpic}[width=1\linewidth]{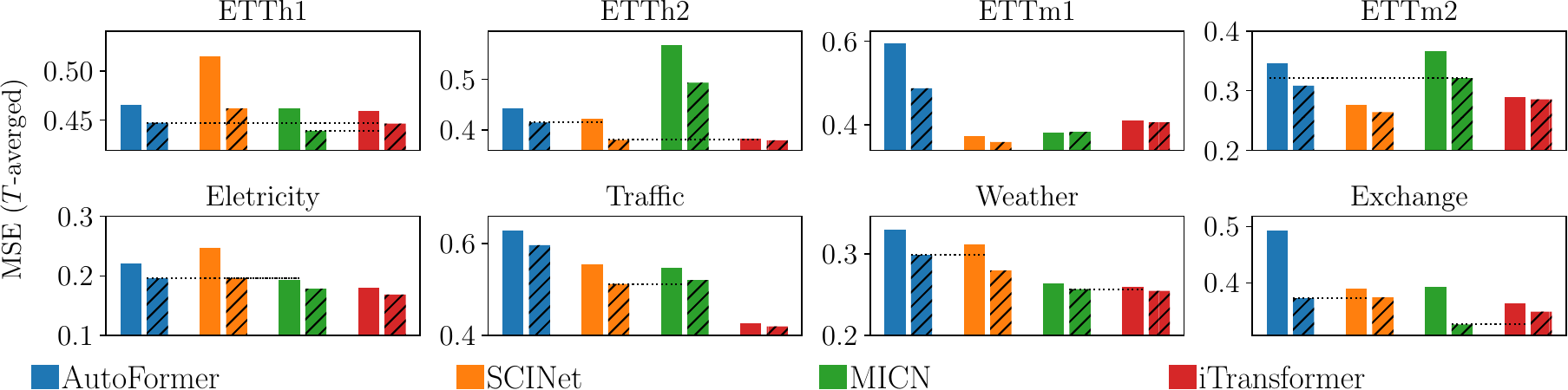}
        \put(14.1, 0.2){\scriptsize{\cite{wu2021autoformer} (NeurIPS2021)}}
        \put(37.3, 0.3){\scriptsize{\cite{liu2022scinet} (NeurIPS2022)}}
        \put(59.7, 0.3){\scriptsize{\cite{wang2023micn} (ICLR2023)}}
        \put(87.2, 0.2){\scriptsize{\cite{liu2024itransformer} (ICLR2024)}}
    \end{overpic}
    \vspace{-1.5em}
    \caption{
        Performance of different models
        with (right striped bars) and without (left color bars) dominant shuffle.
        The horizontal dotted lines demonstrate how dominant shuffle helps
        one model outperforms a more advanced model.
    }\label{fig:sota}
\end{figure}
The results in~\cref{fig:sota} clearly demonstrate that
our method consistently reduces the prediction error
for all the cases.
In some cases,
dominant shuffle surpasses even a highly sophisticated model.
For example, on the ETTh1 dataset,
our approach significantly improves the performance of
AutoFormer~\cite{wu2021autoformer} and MICN~\cite{wang2023micn},
and helps them outperform the latest iTransformer~\cite{liu2024itransformer} model.
On the Exchange and Weather dataset,
our approach enables AutoFormer to outperform SCINet~\cite{liu2022scinet}
and assists MICN~\cite{wang2023micn} in surpassing iTransformer~\cite{liu2024itransformer}.
The results in~\cref{fig:sota} clearly demonstrate 
the significant improvements achieved by our method.

\subsection{Comparisons With Other Data Augmentations}\label{sec:exp-compare-with-other-da}
We compared different data augmentation methods
on various datasets and baseline models under short-term and long-term
protocols.
~\cref{fig:real-perf} demonstrates the relative improvements (\%)
of various augmentation methods over the baseline.
~\cref{tab:short-term,tab:long-term1,tab:long-term2}
summarize the average performance of 5 runs
with distinct random seeds,
and the standard deviations of different runs can be found in~\cref{appendix:perf-std}.
The best values in each colume are highlighted with color.
Example predictions can be found in
~\cref{fig:example-predictions} in the ~\cref{appendix:results}.

We first compared different data augmentation methods
for long-term prediction.
~\cref{tab:long-term1} summarizes the mean squared errors (MSE)
on ETT datasets and
~\cref{tab:long-term2} summarizes the MSE
on Weather, Electricity, and Exchange-rate datasets.
Limited by the space,
we only reported the results of six subsets 
(ETTh1, ETTh2, ETTm1, Electricity, Weather, and Exchange rate)
in~\cref{tab:long-term1,tab:long-term2},
and the results of the other two subsets (ETTm2 and Traffic)
can be found in ~\cref{appendix:results}.
We also merged 
the results of FreqMix and FreqMask by selecting the superior one in each case.
The merged results are denoted as `MixMask'.

\begin{table}[!htb]
    \centering
    \def\arraystretch{0.8}
    \addtolength{\tabcolsep}{-0.30em}
    \resizebox{1\linewidth}{!}{
        \begin{tabular}{c@{\hspace{0.2em}}c|cccc|cccc|cccc}
        \Xhline{2\arrayrulewidth}
        \multicolumn{2}{c}{\multirow{2}{*}{Method}} & \multicolumn{4}{c}{ETTh1} & \multicolumn{4}{c}{ETTh2} & \multicolumn{4}{c}{ETTm1} \\ 
        & & 96 & 192 & 336 & 720 & 96 & 192 & 336 & 720 & 96 & 192 & 336 & 720 \\
        \hline
        \input{tables/long_term_6models_eth3.tex} \\
        \Xhline{2\arrayrulewidth}
    \end{tabular}
    }
    \caption{
        MSE of the long-term prediction
        on the ETT~\cite{zhou2021informer} datasets.
        The best values are marked with colors.
    }\label{tab:long-term1}
\end{table}

As demonstrated in~\cref{tab:long-term1,tab:long-term2},
our method consistently
improves the baseline on \zuojie{96\%} of the cases,
while other augmentation methods, e.g. FreqMix, outperform
the baseline for around \zuojie{87\%} of the cases.
\begin{table}[!htb]
    \centering
    \def\arraystretch{0.8}
    \addtolength{\tabcolsep}{-0.30em}
    \resizebox{1\linewidth}{!}{
        \begin{tabular}{c@{\hspace{0.2em}}c|cccc|cccc|cccc}
            \Xhline{2\arrayrulewidth}
            \multicolumn{2}{c}{\multirow{2}{*}{Method}} & \multicolumn{4}{c}{Electricity} & \multicolumn{4}{c}{Weather} & \multicolumn{4}{c}{Exchange Rate} \\
            & & 96 & 192 & 336 & 720 & 96 & 192 & 336 & 720 & 96 & 192 & 336 & 720 \\
            \Xhline{2\arrayrulewidth}
            \input{tables/long_term_6models_weather3.tex} \\
        \Xhline{2\arrayrulewidth}
\end{tabular}
}
\caption{
MSE of the long-term prediction
on the Weather, Electricity, and Exchange Rate\cite{wu2021autoformer} datasets.
The best values are marked with colors.
}\label{tab:long-term2}
\end{table}
Our method also
outperforms other augmentation methods
on more than \zuojie{77\%} of the cases.
Moreover, our method achieves larger relative improvements
as the prediction length $T$ increased,
highlighting its strong capacity in long-term predictions.
~\cref{tab:short-term}
summarizes the MSE of short-term prediction
using the iTransformer~\cite{liu2024itransformer}
model on the PEMS datasets~\cite{chen2001freeway}.
The prediction errors are generally lower
than the errors in long-term prediction.
Our method outperforms other augmentations in most cases, although the improvements are marginal compared to long-term prediction.
This is because short-term prediction is relatively easy,
and the performance has already reached saturation.

\begin{table}[!htb]
    \centering
    \addtolength{\tabcolsep}{-0.2em}
    \resizebox{0.95\linewidth}{!}{
        \begin{tabular}{l|cccc|cccc|cccc}
        \Xhline{2\arrayrulewidth}
        \input{tables/short_term_1model.tex} \\
        \hline
        \Xhline{2\arrayrulewidth}
        \end{tabular}
    }
    \caption{
        Short-term prediction
        using the iTransformer~\cite{liu2024itransformer}
        on the PEMS datasets~\cite{chen2001freeway}.
    }\label{tab:short-term}
\end{table}

\subsection{Ablation Study}\label{sec:ablation}
Our method includes a hyper-parameter $k$
and two unique designs:
1) perturb the dominant frequencies and
2) shuffle the dominant frequency components.
We conducted ablation studies to investigate
the impact of hyperparameters and to justify our design choices.

\subsubsection{Number of Dominant Frequencies}
\begin{figure}[!htb]
    \begin{overpic}[width=1\linewidth]{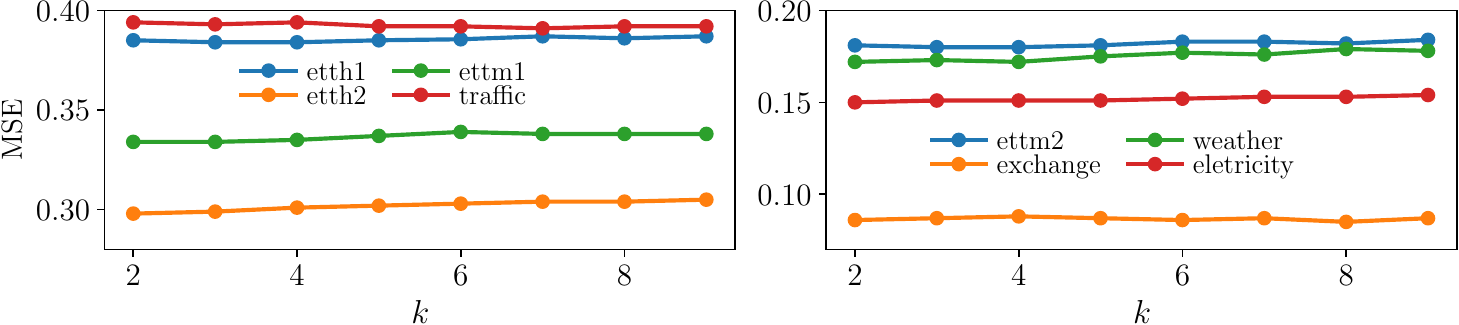}
    \end{overpic}
    \vspace{-2em}
    \caption{
        Mean-squared errors with various $k$ values
        on four datasets under the predict-96 setting.
        Our method is stable against $k$, and the performance varies slightly.
    }\label{fig:ablate-k}
\end{figure}
The only hyper-parameter in our method is
the number of dominant frequencies $k$.
We evaluated the performance using various $k$ values
with iTransformer~\cite{liu2024itransformer}.
The results in~\cref{fig:ablate-k}
reveal that our method is stable against different
$k$ values.

\subsubsection{Shuffle the Dominant Frequencies}
In this experiment, we compared
the combination of different perturbation strategies
and operations.

We first
compared 
perturbing different frequency proportions
including dominant frequencies,
minor frequencies,
and the full spectrum.
The results in~\cref{tab:ablate-dom}
clearly indicate that perturbing the dominant frequencies significantly outperforms other options,
while perturbing the minor frequencies yields the worst performance.
~\cref{tab:ablate-shuf}
compares different perturbation operations
including masking~\cite{chen2023fraug},
adding noise~\cite{gao2020robusttad,lim2021time},
randomization,
and shuffling (ours).
Shuffle consistently surpasses other
operations in most of the cases.

\begin{table}[!htb]
    \centering
    \def\arraystretch{1.1}
    \addtolength{\tabcolsep}{-0.24em}
    \resizebox{0.95\linewidth}{!}{
        \begin{tabular}{ccc|cccc|cccc|cccc}
        \Xhline{2\arrayrulewidth}
            \input{tables/ablate_dom.tex} \\
        \hline
        \Xhline{2\arrayrulewidth}
        \end{tabular}
    }
    \caption{
    Comparison of perturbing 
    different spectrum (full, minor, and dominant) using shuffle
    and random mask.
    Perturbing the dominant frequencies performs
    significantly better than perturbing other frequencies.
    And shuffle is also more effective than random mask.
    }\label{tab:ablate-dom}
\end{table}

\begin{table}[!htb]
\centering
\def\arraystretch{1.1}
\addtolength{\tabcolsep}{-0.24em}
\resizebox{0.95\linewidth}{!}{
    \begin{tabular}{cc|cccc|cccc|cccc}
    \Xhline{2\arrayrulewidth}
        \input{tables/ablate_shuf.tex} \\
    \hline
    \Xhline{2\arrayrulewidth}
    \end{tabular}
}
\caption{
Comparison of different dominant frequency perturbations.
Shuffle outperforms other alternatives with clear margins.
}\label{tab:ablate-shuf}
\end{table}

The results in~\cref{tab:ablate-dom,tab:ablate-shuf}
justified the design decisions in \emph{dominant shuffle}
and confirm that both perturbing dominant frequencies 
and the shuffle operation is superior to other alternatives.
More details about the experiments,
including how we defined minor frequencies
and we implemented mask, noise, and randomization
perturbations can be found in~\cref{appd-impl-details}.

\subsubsection{Different Augmentation Sizes}\label{sec:aug-size}
In prior experiments, we explored data augmentation that doubled the original datasets.
In this experiment, we assessed the performance of various augmentation sizes.
The performance with a larger augmentation size reflects the domain gap between augmented and original data.
A larger augmentation size indicates more augmented samples in the training set.
If these augmented samples are out of distribution compared to the original data, 
larger augmentation sizes could lead to degraded performance due to a training/test gap.
\begin{figure}[!htb]
    \begin{overpic}[width=1\linewidth]{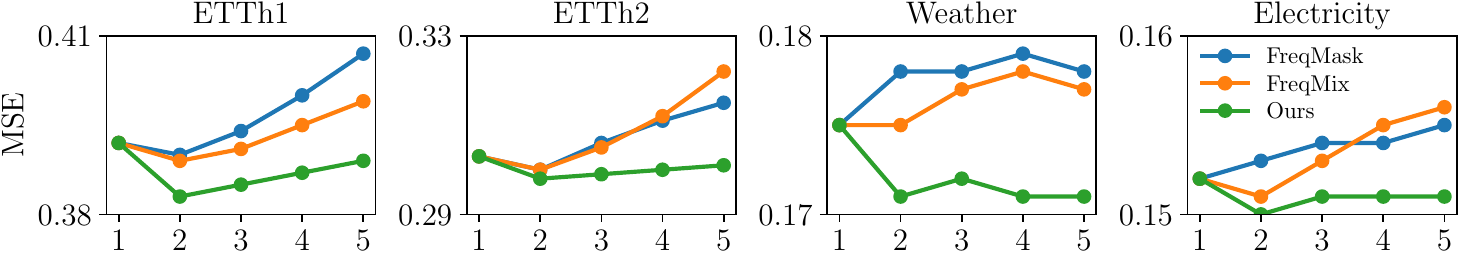}
        \put(45, -1.6){\footnotesize{{\color{gray}{augmentation size}}}}
    \end{overpic}
    \caption{
        MSE with different augmentation sizes using iTransformer~\cite{liu2024itransformer}.
        An augmentation size of two,
        which was used in previous experiments,
        achieves the best results in most cases.
        Our method is more resistant to larger augmentation sizes,
        indicating the improved augmented-original gap.
    }\label{fig:aug-size}
\end{figure}

As shown in~\cref{fig:aug-size},
the performance of FreqMix and FreqMask declines significantly after an augmentation size of two.
This is due to the domain gap between augmented and original data.
Our method is slightly impacted by augmentation size,
and even benefits from larger augmentation sizes on the Weather dataset.
The results in~\cref{fig:aug-size} reveal a smaller augmented-original gap of our method.

\section{Conclusion}\label{sec:conclusion}
We proposed the \emph{dominant shuffle}, 
a simple yet highly effective data augmentation technique for time series prediction.
Our method mitigates the domain gap between
augmented and original data
by limiting the perturbation to dominant frequencies,
and uses shuffles to avoid external noises.
Although being simple and effective, our method is proposed
primarily based on heuristics and lacks theoretical explainability.
Instead of theoretical justifications,
we conducted extensive experiments
using a wide range of datasets,
baseline models,
and augmentation methods 
to validate its consistent improvements across various configurations.
Since dominant shuffle introduces significant
perturbation to the original data
and therefore disrupts the sample-wise class labels,
our method  is limited to prediction tasks
and cannot be extended to classification tasks.
Exploring theoretical justifications and principles
of the proposed method
would be a promising future direction that helps better understand it.

\clearpage
\bibliographystyle{plain}
\bibliography{time-series}

\clearpage

\begin{appendices}
\crefalias{section}{appendix}
\crefalias{subsection}{appendix}

\section{More Details}\label{appendix:details}

\subsection{Datasets}\label{appd:details-datasets}
We evaluate the performance of different models and different augmentations for long-term forecasting on
8 well-established datasets, including Weather, Traffic, Electricity, Exchange Rate~\cite{wu2021autoformer}, and ETT datasets
(ETTh1, ETTh2, ETTm1, ETTm2)~\cite{zhou2021informer}. Furthermore, we adopt PEMS~\cite{chen2001freeway} datasets for short-term
forecasting. We detail the descriptions of the dataset in~\cref{tab:data-stat}.

\begin{table}[!htb]
    \centering
    \resizebox{1\linewidth}{!}{
    \def\arraystretch{1.4}
    \begin{tabular}{c|ccccccccc}
        \Xhline{2\arrayrulewidth}
        \textbf{Dataset} & Variates & Prediction length ($T$) & Total Length (Train:Validation:Test) & Frequency & Information \\
        \hline
        ETTh1,ETTh2& 7&\{96,192,336,720\} &(8545, 2,881, 2,881)&Hourly&Temperature\\
        \hline
        ETTm1,ETTm2& 7&\{96, 192, 336, 720\}&(34465, 11521, 11521)& 15min & Temperature \\
        \hline
        Exchange& 8&\{96, 192, 336, 720\} &(5120, 665, 1422)&Daily&Economy\\
        \hline
        Weather&21&\{96,192,336,720\} & (36792, 5271, 10540)&10min&Weather\\
        \hline
        ECL&321&\{96,192, 336, 720\} & (18317, 2633, 5261)&Hourly&Electricity\\
        \hline
       Traffic&862&\{96, 192, 336, 720\} & (12185, 1757, 3509)&Hourly&Transportation\\
        \hline
        PEMS03&358&\{12, 24, 36, 48\} & (15617, 5135, 5135)&5min&Traffic network\\
        \hline
        PEMS04&307&\{12, 24, 36, 48\}&(10172, 3375, 3375)&5min&Traffic network\\
        \hline
        PEMS07&883&\{12, 24, 36, 48\} & (16911, 5622, 5622) &5min &Traffic network\\
        \hline
        PEMS08 &170 & \{12, 24, 36, 48\}&(10690, 3548, 3548)&5min&Traffic network\\
        \hline
        \Xhline{1.5\arrayrulewidth}
    \end{tabular}
    }\caption{
        Statistics of the eight datasets used in our experiments.
    }\label{tab:data-stat}
\end{table}

\subsection{Implementation Details}\label{appd-impl-details}
\subsubsection{Reimplementation other methods}
For ASD, MSB, and upsample, we reproduce them based on the
descriptions in their original paper~\cite{bandara2021improving,forestier2017generating,semenoglou2023data}. For STAug~\cite{zhang2023towards} and MixMask~\cite{chen2023fraug}, we use their official code. For 
Robusttad~\cite{gao2020robusttad}, we reproduce it by adding Gaussian noise to the frequency components of a time series. For FreqAdd~\cite{zhang2022self}, we perturb a single low-frequency component by setting its magnitude to half of the maximum magnitude.
For FreqPool~\cite{chen2023supervised}, we apply it by maximum pooling of the entire spectrum with size=4. For a fair comparison, all frequency-domain methods target both the data-label pair.

\subsubsection{Different perturbations}
In our ablation study, we define minor frequencies as other components except for the frequency components with the top 10 magnitudes. In~\cref{tab:ablate-dom}, Mask on the full spectrum is similar to FrAug~\cite{chen2023fraug}. Mask on dominant frequencies means mask within frequency components with the top 10 magnitudes, 
Mask on minor frequencies is the opposite. In~\cref{tab:ablate-shuf}, Noise means adding Gaussian noise to the selected frequency components. 
For Random, we first get the maximum and minimum magnitude of the selected frequency components and then randomly assigned magnitude within the max-min range.

\section{More Results}\label{appendix:results}

\subsection{Full forecasting results}

~\cref{tab:app-long-term1-full,tab:app-long-term2-full,tab:app-short-term-full} show the full results of the forecasting task.
Specifically, our method improves the performance of iTransformer by 13$\%$ in Electricity when the predicted length is 720, and it improves the performance of Autoformer by 28$\%$ in ETTm1 when the predicted length is 720. Our method also improves the performance of MICN by 18$\%$ in ETTh2 when the predicted length is 192 and the performance of SCINet by 21$\%$ in Electricity when the predicted length is 720. Similarly, our method improves the performance of Lightts by 29$\%$ in ETTh2 when the predicted length is 720 and the performance of TiDE by 24$\%$ in Electricity when the predicted length is 192. It is worth noting that the strong baseline MixMask falls short in Exchange rate, whose main goal is to predict trends. But our method improves the performance of Autoformer by 34$\%$ in Exchange rate when the predicted length is 720, and it improves the performance of Lightts by 37$\%$ in Exchange rate when the predicted length is 96. These results demonstrate the effectiveness of our method for long-term prediction, as it consistently improves the performance of SOTA methods in different datasets.

\begin{table}[!htb]
    \centering
    \def\arraystretch{0.8}
    \addtolength{\tabcolsep}{-0.30em}
    \resizebox{1\linewidth}{!}{
        \begin{tabular}{c@{\hspace{0.2em}}c|cccc|cccc|cccc|cccc}
        \Xhline{2\arrayrulewidth}
        \multicolumn{2}{c}{\multirow{2}{*}{Method}} & \multicolumn{4}{c}{ETTh1} & \multicolumn{4}{c}{ETTh2} & \multicolumn{4}{c}{ETTm1}  & \multicolumn{4}{c}{ETTm2} \\
        & & 96 & 192 & 336 & 720 & 96 & 192 & 336 & 720 & 96 & 192 & 336 & 720 & 96 & 192 & 336 & 720 \\
        \hline
        \input{tables/long_term_6models_full_ett.tex} \\
        \Xhline{2\arrayrulewidth}
    \end{tabular}
    }
    \caption{
        MSE of the long-term prediction
        on the ETT~\cite{zhou2021informer} datasets.
    }\label{tab:app-long-term1-full}
\end{table}

\begin{table}[!htb]
    \centering
    \def\arraystretch{0.8}
    \addtolength{\tabcolsep}{-0.30em}
    \resizebox{1\linewidth}{!}{
        \begin{tabular}{c@{\hspace{0.2em}}c|cccc|cccc|cccc|cccc}
        \Xhline{2\arrayrulewidth}
        \multicolumn{2}{c}{\multirow{2}{*}{Method}} & \multicolumn{4}{c}{Eletricity} & \multicolumn{4}{c}{Weather} & \multicolumn{4}{c}{Exchange Rate}  & \multicolumn{4}{c}{Traffic} \\
        & & 96 & 192 & 336 & 720 & 96 & 192 & 336 & 720 & 96 & 192 & 336 & 720 & 96 & 192 & 336 & 720 \\
        \hline
        \input{tables/long_term_6models_full_other4.tex} \\
        \Xhline{2\arrayrulewidth}
    \end{tabular}
    }
    \caption{
        MSE of the long-term prediction
        on the Electricity, traffic, Weather, and Exchange Rate~\cite{wu2021autoformer} datasets.
    }\label{tab:app-long-term2-full}
\end{table}

\begin{table}[!htb]
    \centering
    \addtolength{\tabcolsep}{-0.2em}
    \resizebox{0.95\linewidth}{!}{
        \begin{tabular}{l|cccc|cccc|cccc|cccc}
        \Xhline{2\arrayrulewidth}
        \input{tables/short_term_1model_full.tex} \\
        \hline
        \Xhline{2\arrayrulewidth}
        \end{tabular}
    }
    \caption{
        MSE of the Short-term prediction
        using the iTransformer~\cite{liu2024itransformer}
        on the PEMS datasets~\cite{chen2001freeway}.
    }\label{tab:app-short-term-full}
\end{table}

\subsection{Example predictions}
\begin{figure}[!hbt]
    \begin{overpic}[width=1\linewidth]{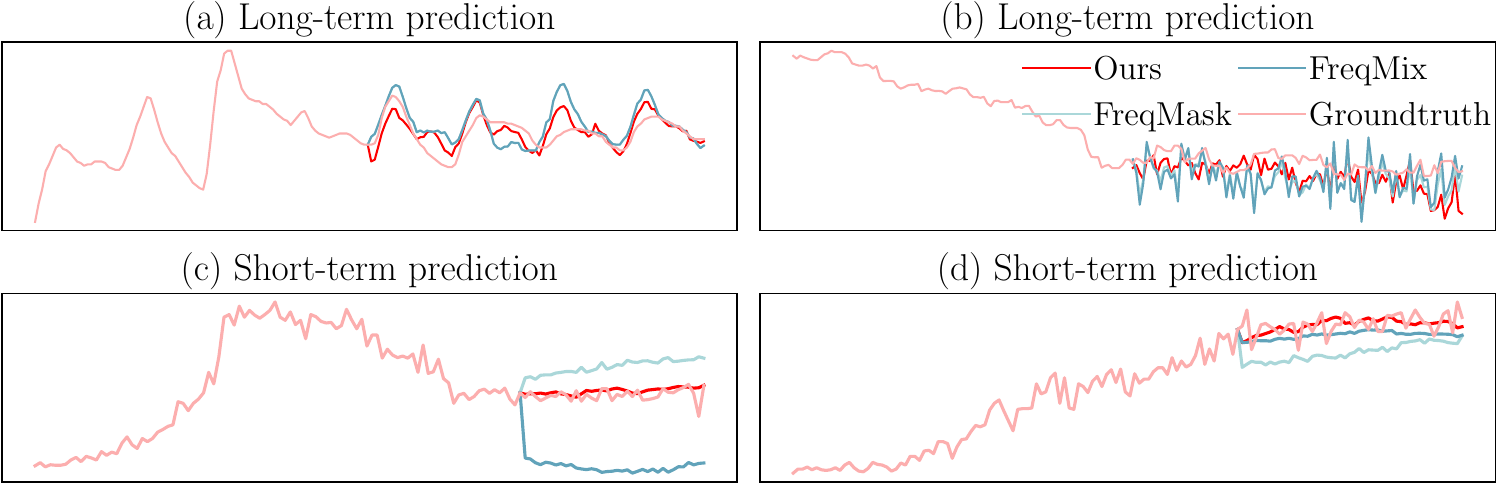}
    \end{overpic}
    \caption{
        Example predictions of different methods under long-term (top)
        and short-term (bottom) protocols.
    }\label{fig:example-predictions}
\end{figure}

We provided 
example prediction results on different datasets in~\cref{fig:example-predictions}

\subsection{Optimal $k$}\label{appd:sec-opt-k}
We provide the optimal $k$ for all long-term prediction datasets using iTranformer~\cite{liu2024itransformer} in~\cref{appendix:tab-opt-k-ett,appendix:tab-opt-k-other4}. As can be seen from the table, our method does not need too much effort to find the optimal parameters.

\begin{table}[!htb]
\centering
\resizebox{0.95\linewidth}{!}{
\begin{tabular}{l|cccc|cccc|cccc|cccc}
\Xhline{2\arrayrulewidth}
\multirow{2}{*}{Hypermeter} & \multicolumn{4}{c}{ETTh1} & \multicolumn{4}{c}{ETTh2} & \multicolumn{4}{c}{ETTm1} & \multicolumn{4}{c}{ETTm2}\\
& 96 & 192 &336 & 720 & 96 & 192 &336 & 720 & 96 & 192 &336 & 720 &96 & 192 &336 & 720 \\
\hline
Optimal $k$ & 4 & 4 & 4 & 4 & 2 & 2 &2 & 4 & 3 & 3 & 2 & 2 &4 & 4 &2 & 4\\
\hline
\end{tabular}
}
\caption{
    The optimal $k$ on ETT datasets
    using the iTransformer~\cite{liu2024itransformer} model.
}\label{appendix:tab-opt-k-ett}
\end{table}

\begin{table}[!htb]
\centering
\resizebox{0.95\linewidth}{!}{
\begin{tabular}{l|cccc|cccc|cccc|cccc}
\Xhline{2\arrayrulewidth}
\multirow{2}{*}{Hypermeter} & \multicolumn{4}{c}{Electricity} & \multicolumn{4}{c}{Traffic} & \multicolumn{4}{c}{Weather} & \multicolumn{4}{c}{Exchange Rate}\\
& 96 & 192 &336 & 720 & 96 & 192 &336 & 720 & 96 & 192 &336 & 720 &96 & 192 &336 & 720 \\
\hline
Optimal $k$ & 2 & 3 & 2 & 2 & 2 & 2 &2 & 2 & 3 & 3 & 2 & 4 &2 & 2 &8 & 8\\
\hline
\end{tabular}
}
\caption{
    The optimal $k$ on Electricity, Traffic, Weather, and Exchange Rate datasets
    using the iTransformer~\cite{liu2024itransformer} model.
}\label{appendix:tab-opt-k-other4}
\end{table}

\subsection{Standard deviations}\label{appendix:perf-std}
~\cref{appendix:tab-std-h1h2,appendix:tab-std-m1m2,appendix:tab-std-eletra,appendix:tab-std-weaex} shows the standard deviations of different runs, indicating the performance of our method is stable.

\begin{table}[!htb]
    \centering
    \addtolength{\tabcolsep}{-0.3em}
    \resizebox{1\linewidth}{!}{
        \begin{tabular}{c@{\hspace{0.2em}}c|cccc|cccc}
        \Xhline{2\arrayrulewidth}
        \multirow{2}{*}{Model} &  & \multicolumn{4}{c}{ETTh1} & \multicolumn{4}{c}{ETTh2}  \\
        & & 96 & 192 & 336 & 720 & 96 & 192 & 336 & 720  \\
        \hline
        \multirow{4}{*}{\makecell{iTrans \\former}} & Baseline& \base{0.392$\pm$0.001} & \base{0.447$\pm$0.002} & \base{0.483$\pm$0.003} & \base{0.516$\pm$0.003} & \base{0.303$\pm$0.001} & \base{0.381$\pm$0.000} & \base{0.412$\pm$0.001} & \base{0.434$\pm$0.002}   \\
        & Mask~\cite{chen2023fraug} & 0.390$\pm$0.001 & 0.442$\pm$0.002 & 0.475$\pm$0.001 & 0.503$\pm$0.003 & 0.301$\pm$0.001 & 0.385$\pm$0.003 & 0.414$\pm$0.001 & 0.438$\pm$0.005  \\
        & Mix~\cite{chen2023fraug} & 0.388$\pm$0.002 & 0.440$\pm$0.002 & 0.477$\pm$0.000 & 0.504$\pm$0.004 & 0.301$\pm$0.001 & 0.380$\pm$0.001 & 0.414$\pm$0.001 & 0.434$\pm$0.003 \\
        & Ours& \best{0.383$\pm$0.001} & \best{0.438$\pm$0.001} & \best{0.473$\pm$0.002} & \best{0.492$\pm$0.002} & \best{0.298$\pm$0.002} & \best{0.382$\pm$0.003} &0.411$\pm$0.004 & \best{0.428$\pm$0.001}  \\
        \hline
        \Xhline{2\arrayrulewidth}
    \end{tabular}
    }\caption{
        Error bars on ETTh1 and ETTh2 datasets.}\label{appendix:tab-std-h1h2}
\end{table}

\begin{table}[!htb]
    \centering
    \addtolength{\tabcolsep}{-0.3em}
    \resizebox{1\linewidth}{!}{
        \begin{tabular}{c@{\hspace{0.2em}}c|cccc|cc cc|cccc|cccc}
        \Xhline{2\arrayrulewidth}
        \multirow{2}{*}{Model} &   & \multicolumn{4}{c}{ETTm1} & \multicolumn{4}{c}{ETTm2} \\
        & & 96 & 192 & 336 & 720 & 96 & 192 & 336 & 720  \\
        \hline
        \multirow{4}{*}{\makecell{iTrans \\former}} & Baseline& \base{0.344$\pm$0.002} & \base{0.383$\pm$0.003} & \base{0.421$\pm$0.001} & \base{0.494$\pm$0.003} & \base{0.183$\pm$0.001} & \base{0.251$\pm$0.002} & \base{0.311$\pm$0.001} & \base{0.412$\pm$0.001}  \\
        & Mask~\cite{chen2023fraug} &  0.347$\pm$0.002 & 0.383$\pm$0.005 & 0.420$\pm$0.001 & 0.494$\pm$0.004 & 0.179$\pm$0.003 & 0.251$\pm$0.001 & 0.311$\pm$0.001 & 0.411$\pm$0.002 \\
        & Mix~\cite{chen2023fraug} & 0.334$\pm$0.005 & 0.375$\pm$0.002 & 0.421$\pm$0.000 & \best{0.485$\pm$0.002} & \best{0.178}$\pm$0.002 & 0.248$\pm$0.001 & 0.311$\pm$0.000 & \best{0.407$\pm$0.002}\\
        & Ours& \best{0.332$\pm$0.001} & \best{0.374$\pm$0.001} & 0.424$\pm$0.001 & 0.492$\pm$0.002 & \best{0.178$\pm$0.002} & \best{0.246$\pm$0.001} & \best{0.309$\pm$0.001} & \best{0.409$\pm$0.000}  \\
        \hline
        \Xhline{2\arrayrulewidth}
    \end{tabular}
    }\caption{
        Error bars on ETTm1 and ETTm2 datasets.}\label{appendix:tab-std-m1m2}
\end{table}

\begin{table}[!htb]
    \centering
    \addtolength{\tabcolsep}{-0.3em}
    \resizebox{1\linewidth}{!}{
        \begin{tabular}{c@{\hspace{0.2em}}c|cccc|cc cc|cccc|cccc}
        \Xhline{2\arrayrulewidth}
        \multirow{2}{*}{Model} &   & \multicolumn{4}{c}{Electricity} & \multicolumn{4}{c}{Traffic}  \\
        & & 96 & 192 & 336 & 720 & 96 & 192 & 336 & 720 \\
        \hline
        \multirow{4}{*}{\makecell{iTrans \\former}} & Baseline& \base{0.152$\pm$0.000} & \base{0.159$\pm$0.001} & \base{0.179$\pm$0.003} & \base{0.230$\pm$0.013} & \base{0.399$\pm$0.001} & \base{0.418$\pm$0.000} & \base{0.428$\pm$0.000} & \base{0.463$\pm$0.000}   \\
        & Mask~\cite{chen2023fraug} & 0.153$\pm$0.001 & 0.157$\pm$0.001 & 0.173$\pm$0.001 & 0.208$\pm$0.005 & 0.395$\pm$0.001 & \best{0.401$\pm$0.005} & \best{0.418$\pm$0.001} & 0.450$\pm$0.002  \\
        & Mix~\cite{chen2023fraug} & 0.151$\pm$0.000 & 0.158$\pm$0.001 & 0.173$\pm$0.000 & 0.205$\pm$0.003 & 0.400$\pm$0.003 & 0.414$\pm$0.004 & 0.424$\pm$0.002 & 0.453$\pm$0.003\\
        & Ours& \best{0.150$\pm$0.000} & \best{0.156$\pm$0.001} & \best{0.171$\pm$0.000} & \best{0.199$\pm$0.002} & \best{0.394$\pm$0.000} & 0.412$\pm$0.002 & 0.423$\pm$0.002 & \best{0.448$\pm$0.001} \\
        \hline
        \Xhline{2\arrayrulewidth}
    \end{tabular}
    }\caption{
        Error bars on Electricity and Traffic datasets.}\label{appendix:tab-std-eletra}
\end{table}

\begin{table}[!htb]
    \centering
    \addtolength{\tabcolsep}{-0.3em}
    \resizebox{1\linewidth}{!}{
        \begin{tabular}{c@{\hspace{0.2em}}c|cccc|cc cc|cccc|cccc}
        \Xhline{2\arrayrulewidth}
        \multirow{2}{*}{Model} &   & \multicolumn{4}{c}{Weather} & \multicolumn{4}{c}{Exchange Rate} \\
        & & 96 & 192 & 336 & 720 & 96 & 192 & 336 & 720  \\
        \hline
        \multirow{4}{*}{\makecell{iTrans \\former}} & Baseline& 0.175$\pm$0.001 & \base{0.224$\pm$0.001} & \base{0.281$\pm$0.000} & \base{0.362$\pm$0.003} & \base{\best{0.086$\pm$0.000}} & \base{0.180$\pm$0.000} & \base{0.335$\pm$0.002} & \base{0.856$\pm$0.004}  \\
        & Mask~\cite{chen2023fraug} & 0.178$\pm$0.001 & 0.228$\pm$0.002 & 0.284$\pm$0.002 & 0.359$\pm$0.001 & 0.090$\pm$0.002 & 0.178$\pm$0.001 & 0.329$\pm$0.006 & 0.845$\pm$0.008 \\
        & Mix~\cite{chen2023fraug} & 0.175$\pm$0.001 & 0.224$\pm$0.000 & 0.279$\pm$0.000 & 0.354$\pm$0.000 & 0.089$\pm$0.001 & 0.178$\pm$0.001 & 0.328$\pm$0.006 & 0.868$\pm$0.008\\
        & Ours&  \best{0.171$\pm$0.001} & \best{0.221$\pm$0.000} & \best{0.276$\pm$0.000} & \best{0.351$\pm$0.002} & \best{0.086$\pm$0.001} & \best{0.176$\pm$0.001} & \best{0.313$\pm$0.006} & \best{0.821$\pm$0.003}\\
        \hline
        \Xhline{2\arrayrulewidth}
    \end{tabular}
    }\caption{
        Error bars on Weather and Exchange Rate datasets.}\label{appendix:tab-std-weaex}
\end{table}

\definecolor{codegreen}{rgb}{0,0.6,0}
\definecolor{codegray}{rgb}{0.5,0.5,0.5}
\definecolor{codepurple}{rgb}{0.58,0,0.82}
\definecolor{backcolour}{rgb}{0.95,0.95,0.92}

\lstdefinestyle{mystyle}{
    backgroundcolor=\color{backcolour},   
    commentstyle=\color{codegreen},
    keywordstyle=\color{magenta},
    numberstyle=\tiny\color{codegray},
    stringstyle=\color{codepurple},
    basicstyle=\ttfamily\footnotesize,
    breakatwhitespace=false,         
    breaklines=true,                 
    captionpos=b,                    
    keepspaces=true,                 
    numbers=left,                    
    numbersep=5pt,                  
    showspaces=false,                
    showstringspaces=false,
    showtabs=false,                  
    tabsize=4
}

\end{appendices}

\ifdefined\preprint
\else
\clearpage
\pagebreak
\newpage
\section*{NeurIPS Paper Checklist}

\begin{enumerate}

\item {\bf Claims}
    \item[] Question: Do the main claims made in the abstract and introduction accurately reflect the paper's contributions and scope?
    \item[] Answer: \answerYes{} 
    \item[] Justification: The effectiveness of our method is
                evidenced by the results in~\cref{tab:long-term1,tab:long-term2,tab:short-term}, and ~\cref{fig:real-perf,fig:sota}.
                The claim that our method is more consistent is evidenced by~\cref{fig:aug-size}.
                And the simplicity of our paper is evidenced by the simple implementation in~\cref{listing:impl}

\item {\bf Limitations}
    \item[] Question: Does the paper discuss the limitations of the work performed by the authors?
    \item[] Answer: \answerYes{} 
    \item[] Justification: The conclusion section discussed the limitation of our method.

\item {\bf Theory Assumptions and Proofs}
    \item[] Question: For each theoretical result, does the paper provide the full set of assumptions and a complete (and correct) proof?
    \item[] Answer: \answerNA{} 
    \item[] Justification: We provided an empirical data augmentation method and there is no theoretical
                assumption and proof.

    \item {\bf Experimental Result Reproducibility}
    \item[] Question: Does the paper fully disclose all the information needed to reproduce the main experimental results of the paper to the extent that it affects the main claims and/or conclusions of the paper (regardless of whether the code and data are provided or not)?
    \item[] Answer: \answerYes{} 
    \item[] Justification: Our method is simple and easy to reproduce.
    We provided all the necessary details of our method.

\item {\bf Open access to data and code}
    \item[] Question: Does the paper provide open access to the data and code, with sufficient instructions to faithfully reproduce the main experimental results, as described in supplemental material?
    \item[] Answer: \answerYes{} 
    \item[] Justification: We provided the code of our DA method in~\cref{listing:impl}.

\item {\bf Experimental Setting/Details}
    \item[] Question: Does the paper specify all the training and test details (e.g., data splits, hyperparameters, how they were chosen, type of optimizer, etc.) necessary to understand the results?
    \item[] Answer: \answerYes{}
    \item[] Justification: We provided all these details in~\cref{sec:impl-details}.

\item {\bf Experiment Statistical Significance}
    \item[] Question: Does the paper report error bars suitably and correctly defined or other appropriate information about the statistical significance of the experiments?
    \item[] Answer: \answerNo{}
    \item[] Justification: Our comparisons were made on large-scale datasets and following the common practices,
                we measured the mean squared error (MSE) of different method without statistical testing.

\item {\bf Experiments Compute Resources}
    \item[] Question: For each experiment, does the paper provide sufficient information on the computer resources (type of compute workers, memory, time of execution) needed to reproduce the experiments?
    \item[] Answer: \answerYes{}
    \item[] Justification: These details are provided in~\cref{sec:impl-details}.
    
\item {\bf Code Of Ethics}
    \item[] Question: Does the research conducted in the paper conform, in every respect, with the NeurIPS Code of Ethics \url{https://neurips.cc/public/EthicsGuidelines}?
    \item[] Answer: \answerYes{} 
    \item[] Justification: We confirmed with the NeurIPS Code of Ethics.

\item {\bf Broader Impacts}
    \item[] Question: Does the paper discuss both potential positive societal impacts and negative societal impacts of the work performed?
    \item[] Answer: \answerNo{} 
    \item[] Justification: The paper does not discuss potential societal impacts as it does not involve human, animal data, nor have any connection with any social impact.
    
\item {\bf Safeguards}
    \item[] Question: Does the paper describe safeguards that have been put in place for responsible release of data or models that have a high risk for misuse (e.g., pretrained language models, image generators, or scraped datasets)?
    \item[] Answer: \answerNo{} 
    \item[] Justification: Our paper has nothing to do with image or language generative models, or scraped datasets.

\item {\bf Licenses for existing assets}
    \item[] Question: Are the creators or original owners of assets (e.g., code, data, models), used in the paper, properly credited and are the license and terms of use explicitly mentioned and properly respected?
    \item[] Answer: \answerNo{} 
    \item[] Justification: We used public datasets without a license.

\item {\bf New Assets}
    \item[] Question: Are new assets introduced in the paper well documented and is the documentation provided alongside the assets?
    \item[] Answer: \answerNo{} 
    \item[] Justification: We are not introducing new datasets nor assets.

\item {\bf Crowdsourcing and Research with Human Subjects}
    \item[] Question: For crowdsourcing experiments and research with human subjects, does the paper include the full text of instructions given to participants and screenshots, if applicable, as well as details about compensation (if any)? 
    \item[] Answer: \answerNA{} 
    \item[] Justification: Our experiments do not include any human subjects.

\item {\bf Institutional Review Board (IRB) Approvals or Equivalent for Research with Human Subjects}
    \item[] Question: Does the paper describe potential risks incurred by study participants, whether such risks were disclosed to the subjects, and whether Institutional Review Board (IRB) approvals (or an equivalent approval/review based on the requirements of your country or institution) were obtained?
    \item[] Answer: \answerNA{}
    \item[] Justification: Our experiments do not include any human subjects.
\end{enumerate}
\fi
\end{document}

%% file: abstract1.tex
%
Recent studies have suggested
frequency-domain Data augmentation (DA)  is effective for time series prediction.
Existing frequency-domain augmentations disturb the original
data with various full-spectrum noises,
leading to excess domain gap between augmented and original data.
Although impressive performance has been achieved
in certain cases,
frequency-domain DA has yet
to be generalized to time series prediction datasets.
In this paper, 
we found that frequency-domain augmentations can be significantly
improved by two modifications that limit the perturbations.
First,
we found that
limiting the perturbation to only dominant frequencies
significantly outperforms full-spectrum perturbations.
Dominant frequencies represent
the main periodicity and trends of the signal
and are more important than other frequencies.
Second, we found that
simply shuffling the dominant frequency components
is superior over sophisticated 
designed random perturbations.
Shuffle rearranges the original components
(magnitudes and phases)
and limits the external noise.
With these two modifications, we proposed \emph{dominant shuffle},
a simple yet effective data augmentation for time series prediction.
Our method is very simple yet powerful and can be implemented with just a few lines of code.
Extensive experiments with eight datasets 
and six popular time series models
demonstrate that
our method consistently improves 
the baseline performance under various settings
and significantly outperforms other DA methods.

%% file: introduction/nips.tex
Time-series prediction aims to forecast
multivariate future values
based on historical observations.
It is a long-standing problem with various applications
in electricity pricing,
weather forecast,
traffic prediction~\cite{lim2021time,zhou2021informer}.
Recently, impressive results have been achieved by using
various deep learning architectures,
e.g. recurrent neural networks (RNNs)
~\cite{rangapuram2018deep,salinas2020deepar,ma2020knowledge},
Transformers~\cite{zhou2021informer,wu2021autoformer,zhou2022fedformer,liu2024itransformer},
and temporal convolutional networks (TCNs)
~\cite{wang2023micn,liu2022scinet,wu2023timesnet}.
Neural networks require a large volume of training data to 
effectively fit their numerous parameters.
Unfortunately, time-series data acquired
from real-world sensors are often limited in
many time-series applications.
The patterns of the time series heavily depend on specific dynamic system
that generates the data
and other data sources are not applicable~\cite{chen2023fraug,semenoglou2023data}.

To mitigate the impact of
insufficient data in time series analysis,
%
%
several data augmentation techniques have been explored
~\cite{ijcai2021p631,forestier2017generating,bandara2021improving,semenoglou2023data,chen2023fraug,zhang2022self,chen2023supervised,gao2020robusttad,zhang2023towards,qian2022makes,um2017data,le2016data,steven2018feature,lim2018doping,nam2020data,lim2021time}.
Most of these data augmentation techniques in time series analysis focus on classification
~\cite{qian2022makes,um2017data,le2016data,steven2018feature,nam2020data,lim2021time,zhang2022self,chen2023supervised}
and anomaly detection~\cite{lim2018doping,lim2021time,gao2020robusttad}.
These augmentations alter the time series sequences while preserving the class labels.
%
%
However, the prediction task requires more fine-grained temporal information
to accurately estimate future dynamics~\cite{zhang2023towards,chen2023fraug}.
These perturbations designed for classification
can disrupt the data-label coherence and 
lead to performance degradation~\cite{zhang2023towards,chen2023fraug}.
%

Coherence is a key factor
to effective data augmentation~\cite{ijcai2021p631,zhang2023towards,sun2023instaboost}.
It measures 
the semantic connection between the augmented data and the label.
These augmentations designed for classification
often struggle with prediction tasks,
due to unilateral perturbations that disrupt the
data-label coherence.
Recently,
to mitigate the data-label coherence,
Chen et. al~\cite{chen2023fraug} proposed to simultaneously
perturb the data (historical sequence)
and labels (future sequences) in the frequency domain.
Unlike common data augmentations 
that introduce slight perturbations only to the data 
while keeping the labels unchanged, 
this approach enables more radical perturbations, 
such as frequency mix (FreqMix) and frequency mask (FreqMask),
to be applied without severely disrupting the data-label coherence.
And the method indeed generates new data-label pairs
that are significantly different from the originals.

%
%
%
However, the full-spectrum perturbations
in FreqMask and FreqMix
introduce external randomization
and reduce the domain gap between the augmented and original data.
This can lead to unstable and suboptimal results
on some benchmarks,
especially with a larger amount of augmented samples.
%
%
As shown in~\cref{fig:aug-size},
the performance of FrAug~\cite{chen2023fraug}
degrades significantly with the rising number of augmented
samples,
which demonstrates that the augmented samples
are out-of-distribution with the original samples.

In this paper, to reduce the domain gap between the augmented and original
data, we propose
to limit the perturbation
and randomization in data augmentation.
First, we limit the perturbation to specific frequencies
instead of full-spectrum perturbation.
Several recent studies have pointed out
that a few frequency components
are dominating the periodicity and main trends of the time series.
And other Frequencies correspond to minor trends or noise
~\cite{wu2023timesnet,zhou2022fedformer,zhou2022film}.
Following~\cite{wu2023timesnet},
we perturb top-$k$ frequencies with highest magnitudes.
Second, to avoid excess external noise,
we use random shuffle for perturbation.
Shuffle rearranges existing components without introducing
any external randomness.
%

\begin{figure}
    \begin{overpic}[width=1\linewidth]{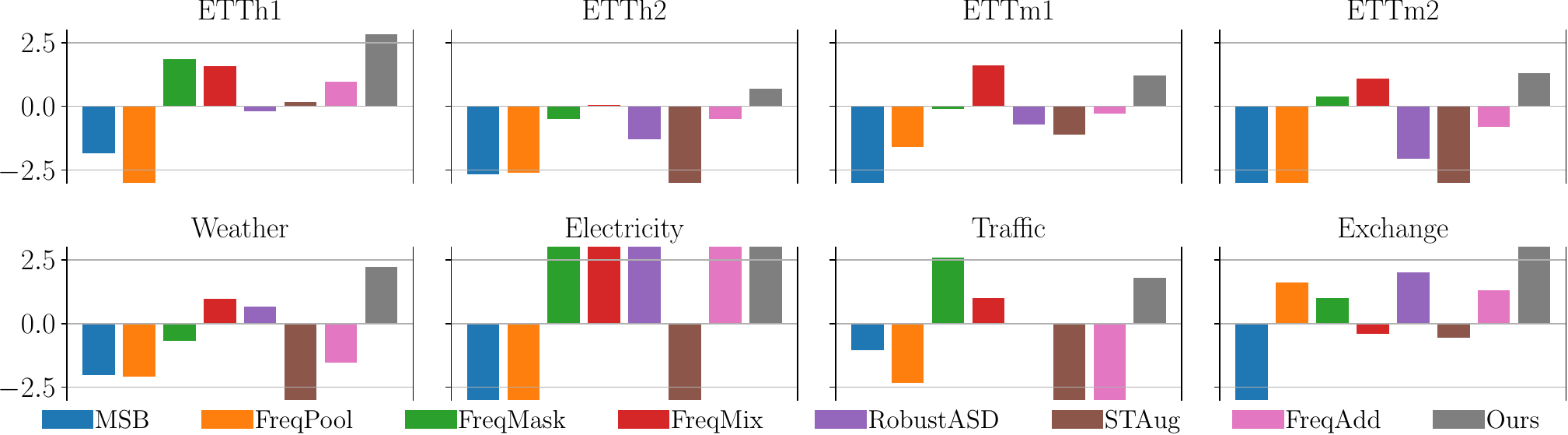}
        \put(9.0,0.375){\footnotesize{~\cite{bandara2021improving}}}
        \put(22.0,0.375){\footnotesize{~\cite{chen2023supervised}}}
        \put(35.5,0.375){\footnotesize{~\cite{chen2023fraug}}}
        \put(48,0.375){\footnotesize{~\cite{chen2023fraug}}}
        \put(62.9,0.375){\footnotesize{~\cite{gao2020robusttad}}}
        \put(74.5,0.372){\footnotesize{~\cite{zhang2023towards}}}
        \put(87.5,0.375){\footnotesize{~\cite{zhang2022self}}}
    \end{overpic}
    \vspace{-1.5em}
    \caption{
        Relative improvements (\%) of various data augmentations
        over the baseline on eight datasets
        using the state-of-the-art iTransformer~\cite{liu2024itransformer} model.
        Zero corresponds to the original model without any data augmentation.
        Our method consistently improves the baseline on all the datasets
        and outperforms other augmentations in most cases.
        The improvements are based on the average performance
        of four prediction lengths: 96, 192, 336, and 720.
    }\label{fig:real-perf}
\end{figure}

%
%
Extensive comparisons were made among nine different data augmentation methods
on eight public datasets using six state-of-the-art time-series prediction network architectures.
%
%
These comparisons demonstrate that,
despite its simplicity,
our method significantly outperforms other competitors
by a substantial margin.
As shown in~\cref{fig:real-perf},
our method consistently improves the performance across various datasets,
and outperforms other augmentations in most cases.

Comprehensive ablation studies demonstrate that perturbing dominant frequencies yields 
significantly better performance than various full-spectrum perturbations.
And shuffle is proven to be superior to other randomization techniques.
Besides, our augmentation demonstrates improved augmented-original gap
over other augmentations,
as indicated by higher performance with an increased number of augmented samples (~\cref{fig:aug-size}).

%
%


%% file: related-work/nips.tex
In the last decade, deep learning has emerged as
a powerful tool in time-series prediction
and has shown superior performance over
traditional statistical methods
such as ARIMA and Exponential Smoothing~\cite{mckenzie1984general}.
A rich line of studies has introduced various deep-learning architectures,
including
recurrent neural networks (RNNs)
~\cite{rangapuram2018deep,salinas2020deepar,ma2020knowledge},
temporal convolution neural networks (TCNs)~\cite{wang2023micn,liu2022scinet,wu2023timesnet},
and Transformers~\cite{wu2021autoformer,ni2023basisformer,nie2023a,liu2024itransformer,zhou2022fedformer}.
These models learn to predict the future
from large volumes of historical data.

Various data augmentations have been proposed for time series data
and many of these techniques were proposed for the classification
tasks~\cite{ijcai2021p631,qian2022makes,um2017data,le2016data,steven2018feature,nam2020data,lim2021time,zhang2022self,chen2023supervised}.
Many of these methods regard time series data as one-dimensional
image and borrowed data augmentations,
e.g. cropping~\cite{le2016data,cui2016multi}
flipping~\cite{ijcai2021p631},
and noise injection~\cite{wen2019time},
from computer vision.
Window warping~\cite{ijcai2021p631}
is a time series-specific data augmentation
that upsamples (or downsamples) a random range of the time series
while keeping other time ranges unchanged.

In addition to time-domain augmentations,
there are also methods that perturb the original data in the frequency
domain.
Gao~\cite{gao2020robusttad} proposed to add noise on both magnitude and phase in the frequency domain. Zhang~\cite{zhang2022self} proposed to add single or multiple
frequency components in the first half of the frequency spectrum. Chen~\cite{chen2023supervised} proposed to perform pooling or smoothing operations in the frequency domain.

While most of the augmentations focus on the classification tasks, a few methods for forecasting task
have also been explored. 
Bandara~\cite{bandara2021improving} introduces two 
DA methods for forecasting
: (i) Average selected with distance (ASD), which generates augmented 
time series using the weighted sum of multiple time series, and the weights 
are determined by the dynamic time warping (DTW) distance\cite{forestier2017generating}; (ii) Moving block bootstrapping 
(MBB) generates augmented data by manipulating the residual part of the time series after STL 
Decomposition~\cite{semenoglou2023data} and recombining it with the other 
series.
Zhang~\cite{zhang2023towards}  proposed to simultaneously augment in frequency and time domains.
Recently,
Chen et. al.~\cite{chen2023fraug} proposed to augment both the data (historical sequence) and the label
(future sequence) in the frequency domain to improve the data-label coherence. Although this method generally achieves decent results, full-spectrum randomization imposes a large domain gap between the augmented and the original data, sometimes leading to degraded performance.


%% file: method/nips.tex
\subsection{Time-series Prediction and Frequency Domain Augmentation}
Time-series prediction is a sequence-to-sequence problem
where the model estimates a future multivariate sequence  based on a 
sequence of historical measurements.
Let $x=\{x^1, x^2, ..., x^L\}_{t=1}^{L}\in\mathbb{R}^{L\times D}$
be the historical sequence,
and $y=\{x^{L+1}, x^{L+2}, ..., x^{L+T}\}_{t=L+1}^{L+T}\in\mathbb{R}^{T\times D}$ is the future sequence to be estimated.
$x^t$ is the measurement at timestep $t$
and $D$ is the number of variates.
Next, we will use $x\in{R}^{L\times D}$ and $y\in{R}^{T\times D}$ 
to denote the historical and future sequences.
$x$ and $y$ are the input and output of deep learning models, respectively.



\subsection{Dominant Frequency Shuffle}
Deep neural networks learn the
$x \rightarrow y$ mapping from large volume of $(x, y)$ pairs,
and data augmentation is an efficient way of expanding the training data.
Frequency-domain augmentation is a family of augmentation methods
that perturb time series in the frequency domain.
These methods initially convert  time series
to the frequency domain, 
apply perturbations there, 
and then convert the modified data back to the time domain. 

\begin{figure}
\begin{overpic}[width=1\linewidth]{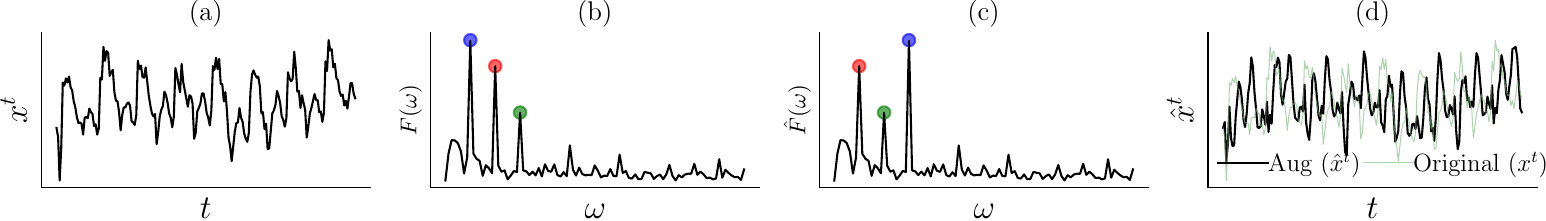}
\end{overpic}
\vspace{-1.5em}
\caption{
    Illustration of shuffling three
    dominant frequencies.
    (a) The original time-series $x^t$.
    (b) and (c) Frequency-domain representations before and after dominant
    shuffle. Color dots represent the shuffle of dominant frequencies.
    (d) Augmented time series with original time series as reference.
}\label{fig:pipeline}
\end{figure}

Following FrAug~\cite{chen2023fraug},
we augmented the concatenation of data and label
to preserve the data-label consistency.
Let $F(\omega) = \mathcal{F}([x, y])$ be the
discrete Fourier transform (DFT)\footnote{
    We used the
    ~\href{https://pytorch.org/docs/stable/generated/torch.fft.rfft.html}{torch.fft.rfft()} and  ~\href{https://pytorch.org/docs/stable/generated/torch.fft.irfft.html}{torch.fft.irfft()}
    for time-to-frequency and inverse conversions.
} of the time-series where
$[x, y]$ denotes the concatenation of data and label.
$F(\omega)$ is the discrete Fourier transform of $[x, y]$.
%
We shuffle
only the dominant frequencies with highest magnitudes ($\lvert F(\omega)\rvert$).
%
%
Let $\hat{F}(\omega)$ be the frequency-domain data
with dominant frequencies shuffled,
$\hat{F}(\omega)$ is then converted back to time domain
using inverse DFT (iDFT): $[\hat{x}, \hat{y}] = \text{iDFT}\big(\hat{F}(\omega)\big)$.
Where $\hat{x}, \hat{y}$ is the augmented
data-label pair.
~\cref{fig:pipeline} illustrates an example
of the process of dominant shuffle with $k=3$.
The prediction models were trained on a combined
training set with both augmented and original data.

%% file: tables/long_term_6models_eth3.tex
\multirow{11}{*}{\rotatebox{90}{iTransformer~\cite{liu2024itransformer}}} & Baseline & \base{0.392} & \base{0.447} & \base{0.483} & \base{0.516} & \base{0.303} & \base{0.381} & \base{0.412} & \base{0.434} & \base{0.344} & \base{0.383} & \base{0.421} & \base{0.494} \\
& ASD~\cite{forestier2017generating} & 0.398 &0.456 &0.483 &0.512 &0.310 &0.388 &0.432 &0.452 &0.340 &0.382 &0.454&0.492\\
& MSB~\cite{bandara2021improving} &0.387&0.460&0.494&0.531&0.309&0.382&0.447&0.433&0.339&0.386&0.467&0.510\\
& Upsample~\cite{semenoglou2023data} & 0.391& 0.445&0.481&0.519&0.305&0.381&0.419&0.430&0.351&0.381&0.432&0.489\\
& FreqAdd~\cite{zhang2022self} & 0.389 & 0.446 & 0.475 & 0.510 & 0.300 & 0.384 & 0.416 & 0.438 & 0.350 & 0.385 & 0.422 & 0.490\\
& FreqPool~\cite{chen2023supervised} &0.433&0.456&0.497&0.532&0.313&0.392&0.415&0.450&0.347&0.392&0.430&0.499\\
& Robusttad~\cite{gao2020robusttad} &0.390&0.445&0.497&0.510&0.312&0.388&0.412&0.439&0.353&0.382&0.421&0.498\\
& STAug~\cite{zhang2023towards} & 0.390 & 0.445 &0.489 & 0.511& 0.323 & 0.428 & 0.486 & 0.483 & 0.339  & 0.383 & \best{0.417} & 0.485 \\
& MixMask~\cite{chen2023fraug} & 0.388 & 0.440 & 0.477 & 0.504 & 0.301 & \best{0.380} & 0.414 & 0.434 & 0.334 & 0.375 & 0.421 & \best{0.485} \\
& Ours& \best{0.383} & \best{0.438} & \best{0.473} & \best{0.492} & \best{0.298} & 0.382 & \best{0.411} & \best{0.428} & \best{0.332} & \best{0.374} & 0.424 & 0.492 \\
\hline
\multirow{11}{*}{\rotatebox{90}{AutoFormer~\cite{wu2021autoformer}}} & Baseline& \base{0.429} & \base{0.440} & \base{0.495} & \base{0.498} & \base{0.381} & \base{0.443} & \base{0.471} & \base{0.475} & \base{0.467} & \base{0.610} & \base{0.529} & \base{0.773} \\
& ASD & 0.450 & 0.485 & 0.523 & 0.556 & 0.370 & 0.465 & 0.476 & 0.503  & 0.480 & 0.620 & 0.502 & 0.633\\
& MSB & 0.462&0.517& 0.612 & 0.579 &0.434 & 0.523 & 0.556 & 0.462 & 0.499 & 0.645 & 0.553 & 0.721 \\
& Upsample & 0.416 & 0.523 & 0.480 & \best{0.482} & 0.353 & 0.460 & 0.455 & 0.509 & 0.498 & 0.630 & 0.512& 0.667\\
& FreqAdd & 0.460 & 0.487 & 0.497 & 0.525 & 0.367 & 0.439 & 0.480 & 0.504 & 0.419 & 0.554 & 0.546 & 0.569 \\
& FreqPool &0.446&0.457&0.523&0.512&0.392&0.442&0.470&0.493&0.479&0.623&0.510&0.754\\
& Robusttad &0.437&0.452&0.492&0.477&0.367&0.497&0.502&0.527&0.432&0.510&0.553&0.623\\
& STAug & 0.429 & 0.478 & 0.505 & 0.506 & 0.354 & 0.443 & 0.496 & 0.495 & 0.415 & 0.581 & 0.588 & 0.693\\
& MixMask & 0.420 & 0.445 & 0.467 & 0.474 & 0.358 & 0.421 & 0.470 & 0.467 & 0.415 & 0.510 & \best{0.491} & 0.588 \\
& Ours& \best{0.409} & \best{0.436} & \best{0.458} & 0.486 & \best{0.335} & \best{0.419} & \best{0.453} & \best{0.452} & \best{0.392} & \best{0.506} & \best{0.491} & \best{0.559} \\
\hline
\multirow{11}{*}{\rotatebox{90}{MICN~\cite{wang2023micn}}} & Baseline& \base{0.384} & \base{0.425} & \base{0.464} & \base{0.574} & \base{0.358} & \base{0.518} & \base{0.566} & \base{0.827} & \base{0.313} & \base{0.360} & \base{0.389} & \base{0.461} \\
& ASD &0.380&0.430&0.472&0.523&0.377&0.539&0.620&0.843&0.315&0.362&0.399&0.457\\
& MSB &0.423&0.423&0.501&0.559&0.402&0.623&0.790&1.126&0.330&0.358&0.402&0.459\\
& Upsample &0.396&0.435&0.463&0.550&0.366&0.500&0.831&0.752&0.339&0.377&0.402&0.475\\
& FreqAdd & 0.390 & 0.430 & 0.477 & 0.643 & 0.370 & 0.521 & 0.626 & 0.975 & 0.316 & 0.360 & 0.407 & 0.478\\
& FreqPool &0.399&0.465&0.473&0.572&0.365&0.553&0.550&0.812&0.336&0.372&0.397&0.466\\
& Robusttad &0.392&0.436&0.491&0.556&0.339&0.529&0.553&0.998&0.339&0.359&0.396&0.472\\
& STAug & 0.374 & 0.429 & 0.489 & 0.608 & 0.413 & 0.760 & 1.330 & 2.608 & 0.313 & 0.360 & 0.418 & 0.483 \\
& MixMask & 0.378 & 0.423 & 0.461 & 0.521 & 0.339 & 0.488 & 0.544 & 0.735 & \best{0.301} & \best{0.352} & 0.401 & \best{0.454} \\
& Ours& \best{0.373} & \best{0.421} & \best{0.452} & \best{0.510} & \best{0.310} & \best{0.427} & \best{0.507} & \best{0.731} & 0.314 & 0.360 & \best{0.387} & 0.470 \\
\hline
\multirow{11}{*}{\rotatebox{90}{SCINet~\cite{liu2022scinet}}} & Baseline& \base{0.485} & \base{0.506} & \base{0.519} & \base{0.552} & \base{0.372} & \base{0.416} & \base{0.429} & \base{0.470} & \base{0.316} & \base{0.353} & \base{0.387} & \base{0.431} \\
& ASD &0.494 & 0.480&0.491&0.559&0.362&0.402&0.432&0.499&0.331&0.367&0.389&0.453\\
& MSB &0.489 & 0.466&0.502&0.547&0.359&0.396&0.458&0.476&0.320&0.351&0.396&0.478\\
& Upsample & 0.471&0.457&0.479&0.541	&0.379&	0.407&	0.403&	0.482&	0.342&	0.386&	0.399	&0.442\\
& FreqAdd & 0.428 & 0.452 & 0.469 & 0.532 & 0.335 & 0.385 & 0.403 & 0.447 & 0.304 & \best{0.338} &0.373&0.421\\
& FreqPool &0.499&0.510&0.557&0.549&0.410&0.453&0.432&0.475&0.331&0.362&0.379&0.432\\
& Robusttad &0.462&0.501&0.498&0.559&0.362&0.431&0.419&0.496&0.331&0.351&0.394&0.438\\
& STAug & 0.457 & 0.500 & 0.524 & 0.534 & 0.538 & 0.636 & 0.681 & 0.648 & 0.319 & 0.357 & 0.389 & 0.445\\
& MixMask & 0.427 & 0.452 & 0.465 & 0.548 & \best{0.335} & 0.377 & 0.400 & 0.438 & \best{0.302} & 0.341 & 0.376 & 0.423 \\
& Ours& \best{0.417} & \best{0.443} & \best{0.461} & \best{0.527} & \best{0.335} & \best{0.375} & \best{0.392} & \best{0.421} & \best{0.302} & \best{0.338} & \best{0.372} & \best{0.420} \\
\hline
\multirow{11}{*}{\rotatebox{90}{TiDE~\cite{das2023longterm}}} & Baseline& \base{0.401} & \base{0.434} & \base{0.521} & \base{0.558} & \base{0.304} & \base{0.350} & \base{0.331} & \base{0.399} & \base{0.311} & \base{0.340} & \base{0.366} & \base{0.420} \\
& ASD &0.417&0.441&0.513&0.556&0.320&0.351&0.367&0.422&0.319&0.341&0.399&0.432\\
& MSB &0.422&0.476&0.529&0.579&0.331&0.379&0.334&0.401&0.302&0.356&0.382&0.451\\
& Upsample &0.431&0.452&0.533&0.604&0.346&0.372&0.350&0.456&0.324&0.339&0.378&0.463\\
& FreqAdd & 0.385 & 0.420 & 0.477 & 0.505 & 0.289 & 0.336 & 0.330 & 0.390 & 0.309 & 0.339 &0.365&0.417\\
& FreqPool &0.423&0.455&0.510&0.592&0.312&.376&0.339&0.397&0.319&0.352&0.397&0.453\\
& Robusttad &0.396&0.432&0.521&0.537&0.331&0.352&0.337&0.398&0.321&0.346&0.382&0.437\\
& STAug & 0.515 & 0.535 & 0.521 & 0.558 & 0.390 & 0.437 & 0.403 & 0.508 & 0.310 & 0.337 & \best{0.364} & 0.417\\
& MixMask & \best{0.385} & 0.420 & 0.478 & 0.507 & 0.289 & 0.339 & 0.330 & 0.391 & 0.299 & 0.332 & 0.367 & 0.416 \\
& Ours& \best{0.385} & \best{0.414} & \best{0.467} & \best{0.498} & \best{0.283} & \best{0.332} & \best{0.324} & \best{0.388} & \best{0.297} & \best{0.328} & 0.365 & \best{0.412} \\
\hline
\multirow{11}{*}{\rotatebox{90}{LightTS~\cite{zhang2022less}}} & Baseline& \base{0.448} & \base{0.444} & \base{0.663} & \base{0.706} & \base{0.369} & \base{0.476} & \base{0.738} & \base{1.165} & \base{0.323} & \base{0.347} & \base{0.428} & \base{0.476} \\
& ASD &0.451&0.476&0.633&0.681&0.392&0.469&0.701&0.998&0.356&0.352&0.441&0.478\\
& MSB &0.467&0.463&0.627&0.652&0.378&0.472&0.652&1.123&0.371&0.349&0.430&0.479\\
& Upsample &0.449&0.472&0.610&0.637&0.401&0.487&0.714&1.245&0.329&0.366&0.453&0.492\\
& FreqAdd & 0.417 & 0.430 & 0.578 & 0.622 & 0.351 & 0.453 & 0.689 & 1.125 & 0.322 &0.352&0.400&0.450\\
& FreqPool &0.463&0.471&0.652&0.690&0.369&0.512&0.723&1.264&0.336&0.351&0.442&0.497\\
& Robusttad &0.445&0.442&0.590&0.654&0.372&0.468&0.699&0.982&0.331&0.352&0.441&0.462\\
& STAug & 0.445 & 0.441 & 0.669 & 0.714 & 0.520 & 0.807 & 2.101 & 2.467 & 0.320 & 0.343 & 0.427 & 0.476\\
& MixMask & 0.417 & 0.429 & 0.575 & 0.620 & 0.337 & 0.426 & 0.643 & 0.993 & \best{0.316} & \best{0.340} & 0.398 & 0.447 \\
& Ours& \best{0.405} & \best{0.423} & \best{0.565} & \best{0.603} & \best{0.335} & \best{0.395} & \best{0.575} & \best{0.827} & 0.322 & \best{0.340} & \best{0.391} & \best{0.440} 

%% file: tables/long_term_6models_weather3.tex
\multirow{11}{*}{\rotatebox{90}{iTransformer~\cite{liu2024itransformer}}} & Baseline& \base{0.152} & \base{0.159} & \base{0.179} & \base{0.230} & \base{0.175} & \base{0.224} & \base{0.281} & \base{0.362} & \base{\best{0.086}} & \base{0.180} & \base{0.335} & \base{0.856}  \\
& ASD~\cite{forestier2017generating} &0.173&	0.179&	0.201	&0.234&0.191 &	0.223	&0.280 &	0.364&	0.088&	0.183&	0.343&	0.872\\
& MSB~\cite{bandara2021improving} &0.182	&0.182 &	0.194&	0.267&0.185	&0.235&	0.284&	0.359 &	0.089	&0.189	&0.359	&0.907\\
& Upsample~\cite{semenoglou2023data} &0.166	&0.188&	0.216	&0.221&0.204	&0.257&	0.291	&0.373&	0.086&	0.180 	&0.338 	&0.834\\
& FreqAdd~\cite{zhang2022self} & \best{0.150} & 0.157 & 0.172 & 0.204 & 0.181 & 0.230 & 0.285 & 0.362 & 0.087 & 0.181 
    &0.333 &  0.837   \\
& FreqPool~\cite{chen2023supervised} &0.169&0.170&0.194&0.237&0.184&0.223&0.279&0.378&0.088&0.183&0.330&0.832\\
& Robusttad~\cite{gao2020robusttad} &0.150&0.157&0.176&0.210&0.172&0.225&0.281&0.357&0.087&0.179&0.329&0.833\\
& STAug~\cite{zhang2023towards} & 0.160 & 0.173 & 0.218 & 0.372 & 0.206 & 0.264 & 0.319 & 0.385 & 0.086 & 0.178 &0.335 & 0.866\\
& MixMask~\cite{chen2023fraug} & 0.151 & 0.158 & 0.173 & 0.205 & 0.175 & 0.224 & 0.279 & 0.354 & 0.089 & 0.178 & 0.328 & 0.845\\
& Ours& \best{0.150} & \best{0.156} & \best{0.171} & \best{0.199} &  \best{0.171} & \best{0.221} & \best{0.276} & \best{0.351} & \best{0.086} & \best{0.176} & \best{0.313} & \best{0.821}\\
\hline
\multirow{11}{*}{\rotatebox{90}{AutoFormer~\cite{wu2021autoformer}}} & Baseline& \base{0.203} & \base{0.208} & \base{0.231} & \base{0.239} & \base{0.241} & \base{0.314} & \base{0.341} & \base{0.425} & \base{0.143} & \base{0.305} & \base{0.470} & \base{1.056}  \\
& ASD &0.247&	0.216	&0.221	&0.235&0.652&	0.392&	0.416	&0.513&	0.141&	0.280 &	0.579	&1.240\\
& MSB &0.237	&0.256&	0.295	&0.236&0.256	&0.379	&0.402&	0.468&	0.156&	0.254	&0.513	&1.339\\
& Upsample &0.201&	0.209	&0.232	&0.268&0.281&	0.294&	0.329	&0.385	&0.141	&0.292	&0.553&	1.295\\
& FreqAdd & 0.193 & 0.197 & 0.212 & 0.225 & 0.255 & 0.323 & 0.370 & 0.419 & 0.143 & 0.369 & 0.716 & 1.173 \\
& FreqPool &0.213&	0.224&	0.234&	0.257&0.237&	0.339	&0.372&	0.446&	0.142&	0.336	&0.532&	1.014\\
& Robusttad &0.230 &	0.242	&0.261	&0.231&0.27&	0.334&	0.351&	0.429	&0.142	&0.309	&0.462	&1.123\\
& STAug & 0.191 & 0.206 & 0.217 & 0.234 & 0.250 & 0.300 & 0.347 & 0.418 & 0.140 & 0.326 & 0.594 & 1.176\\
& MixMask & 0.177 & 0.194 & 0.206 & 0.224 & 0.240 & 0.302 & 0.330 & 0.422 & 0.141 & 0.284 & 0.453 & 0.778 \\
& Ours& \best{0.171} & \best{0.191} & \best{0.203} & \best{0.219}  & \best{0.214} & \best{0.273} & \best{0.327} & \best{0.383} & \best{0.136} & \best{0.243} & \best{0.418} & \best{0.695} \\
\hline
\multirow{11}{*}{\rotatebox{90}{MICN~\cite{wang2023micn}}} & Baseline& \base{0.171} & \base{0.183} & \base{0.198} & \base{0.224} & \base{0.188} & \base{0.241} & \base{0.278} & \base{0.350} & \base{0.091} & \base{0.185} & \base{0.355} & \base{0.941}  \\
& ASD &0.165	&0.174&	0.190	&0.237&0.189&	0.242	&0.276&	0.354	&0.087	&0.175 	&0.337	&1.203\\
& MSB &0.179	&0.182&	0.201	&0.225&0.201&	0.250 &	0.291&	0.365&	0.088	&0.176&	0.360 &	0.995\\
& Upsample &0.182	&0.180 &	0.203	&0.220 &0.193	&0.249	&0.279&	0.372	&0.084&	0.171	&0.313	&\best{0.702}\\
& FreqAdd & 0.160 & 0.169 & 0.182 & 0.199 & 0.180 & 0.234 & 0.282 & 0.350 & 0.087 & 0.174 & 0.349 & 0.923 \\
& FreqPool &0.182&	0.203&	0.241	&0.256&0.192	&0.257	&0.278	&0.351	&0.089	&0.179	&0.394	&0.923\\
& Robusttad &0.179&	0.220& 	0.234&	0.227&0.192	&0.239&	0.292&	0.343&	0.085&	0.179	&0.336&	0.932\\
& STAug & 0.180 & 0.195 & 0.210 & 0.224 & 0.272 & 0.356 & 0.433 & 0.559 & 0.092 & 0.183 & 0.313 & 0.790 \\
& MixMask & 0.159 & \best{0.165} & \best{0.178} & \best{0.195} & 0.185 & 0.239 & 0.281 & 0.344 & 0.086 & 0.174 & 0.337 & 0.796 \\
& Ours& \best{0.157} & 0.168 & \best{0.178} & 0.211 & \best{0.179} & \best{0.232} & \best{0.275} & \best{0.342} & \best{0.084} & \best{0.169} & \best{0.303} & 0.750 \\
\hline
\multirow{11}{*}{\rotatebox{90}{SCINet~\cite{liu2022scinet}}} & Baseline& \base{0.212} & \base{0.237} & \base{0.255} & \base{0.286} & \base{0.229} & \base{0.282} & \base{0.334} & \base{0.402} & \base{0.099} & \base{0.191} & \base{0.356} & \base{0.916}  \\
& ASD &0.229&	0.241&	0.239	&0.282&0.254	&0.276&	0.356&	0.462&	0.095	&0.204&	0.379	&1.230\\
& MSB &0.232&	0.237&	0.228	&0.274&0.279	&0.265	&0.374	&0.454	&0.093	&0.267&	0.402	&0.965\\
& Upsample &0.250	&0.232	&0.271&	0.309&0.243	&0.299&	0.361	&0.431	&0.092&	0.196&	\best{0.311}	&0.932\\
& FreqAdd & 0.176 & 0.195 & 0.212 & 0.237 & 0.208 & 0.258 & 0.309 & 0.385 & 0.092 & 0.186 & 0.343 & 0.920\\
& FreqPool &0.230	&0.221	&0.242	&0.339&0.261	&0.290 &	0.337&	0.456	&0.096	&0.183	&0.551&	0.938\\
& Robusttad &0.189&	0.202	&0.210 	&0.243&0.229&	0.281	&0.331&	0.410 &	0.093	&0.186	&0.334	&0.957\\
& STAug & 0.210&0.239 & 0.282 & 0.411 & 0.277 & 0.329  & 0.372 & 0.435 & 0.098 & 0.191 & 0.342 & 0.931 \\
& MixMask & \best{0.171} & \best{0.188} & 0.204 & 0.230  & 0.205 & 0.250 & 0.310 & \best{0.374} & 0.093 & 0.179 & 0.336 & 0.928 \\
& Ours& 0.172 & \best{0.188} & \best{0.200} & \best{0.225} & \best{0.197} & \best{0.246} & \best{0.299} & 0.379 & \best{0.091} & \best{0.175} & 0.342 & \best{0.890} \\
\hline
\multirow{11}{*}{\rotatebox{90}{TiDE~\cite{das2023longterm}}} & Baseline& \base{0.207} & \base{0.197} & \base{0.211} & \base{0.238} & \base{0.177} & \base{0.220} & \base{0.265} & \base{0.323} & \base{0.093} & \base{0.184} & \base{0.330} & \base{0.860}  \\
& ASD &0.232&0.220&0.231&0.265&0.189&0.221&0.297&0.332&0.095&0.206&0.351&0.962\\
& MSB &0.210&0.219&0.253&0.261&0.199&0.254&0.273&0.339&0.092&0.179&0.358&0.941\\
& Upsample &0.206&0.199&0.223&0.274&0.203&0.267&0.331&0.355&0.091&0.182&0.331&0.852\\
& FreqAdd & 0.150 & 0.163 & 0.177 & 0.209 & \best{0.173} & \best{0.216} & 0.263 & \best{0.322} & \best{0.088} & 0.180 & 0.330 & 0.848 \\
& FreqPool &0.224&0.238&0.233&0.270&0.189&0.224&0.292&0.334&0.092&0.334&0.521&1.124\\
& Robusttad &0.176&0.166&0.182&0.229&0.182&0.231&0.279&0.330&0.099&0.232&0.331&0.924\\
& STAug & 0.230 & 0.210& 0.192 & 0.225 & 0.205 & 0.247 & 0.292 & 0.364 & 0.092 & 0.184 & 0.330 & 0.859 \\
& MixMask & \best{0.143} & 0.155 & \best{0.164} & 0.210 & \best{0.173} & \best{0.216} & 0.263 & 0.323 & 0.089 & 0.180 & 0.329 & 0.861 \\
& Ours& \best{0.143} & \best{0.150} & 0.165 & \best{0.202} & 0.177 & 0.219 & \best{0.261} & \best{0.322} & \best{0.088} & \best{0.179} & \best{0.324} & \best{0.847} \\
\hline
\multirow{11}{*}{\rotatebox{90}{LightTS~\cite{zhang2022less}}} & Baseline& \base{0.210} & \base{0.169} & \base{0.182} & \base{0.212} & \base{0.168} & \base{0.210} & \base{0.260} & \base{0.320} & \base{0.139} & \base{0.252} & \base{0.412} & \base{0.840}  \\
& ASD &0.225	&0.179	&0.198&	0.232&0.179	&0.210	&0.271&	0.321&	0.132	&0.320 &	0.436&	1.036\\
& MSB &0.233	&0.182	&0.204	&0.228&0.170	&0.214&	0.259	&0.332&	0.117&	0.294	&0.502	&0.964\\
& Upsample &0.246 &	0.179	 &0.211 &	0.254 &0.182	 &0.223	 &0.257	 &0.336	 &0.099 &	0.251 &	0.369	 &0.702\\
& FreqAdd & 0.213 & 0.159 & 0.177 & 0.210 & 0.164 & 0.207 & 0.258 & 0.317 & 0.098 & 0.522 & 0.565 & 1.583\\
& FreqPool &0.219&	0.174	&0.197&	0.236&0.193	&0.254	&0.267&	0.339&	0.099	&0.275&	0.394&	0.793\\
& Robusttad &0.212&	0.169&	0.181&	0.223&0.172&	0.223	&0.259	&0.324	&0.092	&0.279	&0.451	&0.796\\
& STAug & 0.224 & 0.267 & 0.294 & 0.351 & 0.214 & 0.263 & 0.382 & 0.371 & 0.096 & \best{0.212} & 
0.380 & 0.690 \\
& MixMask & \best{0.192} & 0.158 & 0.175 & 0.211 & \best{0.163} & 0.206 & 0.257 & 0.318 & 0.099 & 0.384 & 0.518 & 0.774 \\
& Ours& 0.210 & \best{0.156} & \best{0.173} & \best{0.206} & 0.165 & \best{0.205} & \best{0.249} & \best{0.312} & \best{0.088} & 0.243 & \best{0.361} & \best{0.676}

%% file: tables/short_term_1model.tex
\multirow{2}{*}{Methods} & \multicolumn{4}{c}{PEMS03} & \multicolumn{4}{c}{PEMS04} & \multicolumn{4}{c}{PEMS07} \\
& 12 & 24 & 36 & 48 & 12 & 24 & 36 & 48 & 12 & 24 & 36 & 48 \\
\hline
 Baseline& \base{0.070} & \base{0.097} & \base{0.134} & \base{0.164} & \base{0.088} & \base{0.124} & \base{0.160} & \base{0.196} & \base{0.067} & \base{0.097} & \base{0.128} & \base{0.156}  \\
ASD~\cite{forestier2017generating} &0.072&0.096&0.152&0.239&0.098&0.132&0.156&0.190&0.069&0.099&0.154&0.181\\
MSB~\cite{bandara2021improving} &0.096&0.131&0.129&0.214&0.087&0.134&0.167&0.219&0.098&0.096&0.137&0.165\\
Upsample~\cite{semenoglou2023data} &0.069&0.096&0.128&0.179&0.087&0.124&0.158&0.199&0.072&0.099&0.127&0.155\\
FreqAdd~\cite{zhang2022self} &1.036	&0.104	&0.251&	0.362&0.088	&0.125	&0.159&	0.201	&0.067&	0.097	&0.127&	0.155\\
FreqPool~\cite{chen2023supervised} &1.234	&0.178	&0.296&	0.451	&0.099	&0.145	&0.178	&0.226	&0.079	&0.104	&0.152	&0.172\\
Robusttad~\cite{gao2020robusttad} &0.082&	0.098&	0.132&	1.520&	0.089&	0.123	&0.161	&0.195	&0.067&	0.097	&0.129	&0.157\\
STAug~\cite{zhang2023towards} &0.079&	0.112&	0.195&	0.456	&0.087&	0.120 &	0.162	&0.304	&0.066	&0.096	&0.132	&0.165\\
Mask~\cite{chen2023fraug} & 0.443 & 1.205 & 0.233 & 1.510 & 0.086 & 0.119 & 0.158 & 0.346 & \best{0.065} & 0.095 & 0.125 & 0.156 \\
Mix~\cite{chen2023fraug} & 1.018 & 0.097 & 0.877 & 1.501 & \best{0.085} & 0.119 & 0.154 & 0.205 & \best{0.065} & \best{0.094} & 0.134 & 0.152\\
Ours& \best{0.067} & \best{0.095} & \best{0.126} & 0.235 &  \best{0.085} & \best{0.118} & \best{0.149} & \best{0.182} & \best{0.065} & \best{0.094} & \best{0.123} & \best{0.148}

%% file: tables/ablate_dom.tex
& & & \multicolumn{4}{c}{ETTh1} & \multicolumn{4}{c}{ETTm2} & \multicolumn{4}{c}{Weather} \\
& & & 96 & 192 & 336 & 720 & 96 & 192 & 336 & 720 & 96 & 192 & 336 & 720 \\
\hline
\multirow{6}{*}{\rotatebox{90}{iTrans~\cite{liu2024itransformer}}} &
\multirow{3}{*}{\rotatebox{90}{Shuffle}}
& full & 0.391 & 0.447 & 0.486 & 0.509 & 0.182 & 0.247 & 0.311 & \best{0.403} & 0.175 & 0.223 & 0.278 & 0.355 \\
& & min & 0.389 & 0.445 & 0.494& 0.505 & 0.181 & 0.251 & 0.310 & 0.413 & 0.174 & 0.225 & 0.282 & 0.355  \\
& & dom & \best{0.383} & \best{0.438} & \best{0.473} & \best{0.492}
& \best{0.178} & \best{0.246} & \best{0.309} & 0.409 & \best{0.171} & \best{0.221} & \best{0.276} & \best{0.351} \\
\cdashline{2-15}
& \multirow{3}{*}{\rotatebox{90}{Mask}}
&  full & 0.390 & \best{0.442} & \best{0.475} & 0.503 & \best{0.179} &\best{0.251} & 0.311 & 0.411 & 0.178 & 0.228 & 0.284 & 0.359  \\
& &min & 0.389 & 0.444 & 0.487 & \best{0.499} & 0.183 & 0.252 & 0.311 & 0.412 & 0.180 & 0.226 & 0.282 & 0.361  \\
& & dom & \best{0.388} & \best{0.442} & 0.486 & 0.505
& 0.180 & \best{0.251} & \best{0.309} & \best{0.410} & \best{0.173} & \best{0.224} & \best{0.280} & \best{0.356} \\
\hline
\multirow{6}{*}{\rotatebox{90}{MICN~\cite{wang2023micn}}} &
\multirow{3}{*}{\rotatebox{90}{Shuffle}}
& full &0.385 & 0.427 &  0.466 &  0.604 & 0.184 & 0.293 & 0.375  & 0.594 & 0.182 & 0.239 & 0.280 & 0.348 \\
& & min &0.390 & 0.430 & 0.480 & 0.565 & 0.191 & 0.281 & 0.365 & 0.580 & 0.197 & 0.236 & 0.283 & 0.349   \\
& & dom & \best{0.373} & \best{0.421} & \best{0.452} & \best{0.510} & \best{0.174} & \best{0.263} & \best{0.348} & \best{0.502} & \best{0.179} & \best{0.232} & \best{0.275} & \best{0.342} \\
\cdashline{2-15}
& \multirow{3}{*}{\rotatebox{90}{Mask}}
&  full & 0.381 & 0.424 & 0.460 & \best{0.543} & 0.184 & \best{0.265} & 0.353 & 0.510 & 0.190 & \best{0.236} & \best{0.281} & 0.345  \\
& &min & 0.385 & 0.426 & 0.472 & 0.553 & 0.187 & 0.276 & 0.359 & 0.542 & 0.179 & 0.240 & \best{0.281} & 0.344  \\
& & dom & \best{0.377} & \best{0.421} & \best{0.454} & 
\best{0.543} & \best{0.175} & 0.268 & \best{0.337} & \best{0.505} & \best{0.178} & 0.239 & 0.283 & \best{0.342} \\
\hline
\multirow{6}{*}{\rotatebox{90}{Lightts~\cite{zhang2022less}}} &
\multirow{3}{*}{\rotatebox{90}{Shuffle}}
& full & 0.415 & 0.426 & 0.577 & 0.621 & 0.202 & \best{0.235} & 0.325 & 0.445 & \best{0.163} & \best{0.205} & 0.251 & 0.317 \\
& & min & 0.418 & 0.432 & 0.577 & 0.619 & 0.206 & 0.239 & 0.326 & 0.444 & 0.164 & 0.212 & 0.259 & 0.317  \\
& & dom & \best{0.405} & \best{0.423} & \best{0.565} & \best{0.603} & \best{0.195} & 0.245 & \best{0.312} & \best{0.422} & 0.165 & \best{0.205} & \best{0.249} & \best{0.312} \\
\cdashline{2-15}
& \multirow{3}{*}{\rotatebox{90}{Mask}}
&  full & \best{0.418} & 0.432 & \best{0.573} & 0.621 & 0.204 & \best{0.238} & 0.321 & 0.435 & 0.163 & 0.206 & 0.258 & \best{0.317}  \\
& &min & 0.419 & 0.433 & 0.578 & 0.621 & 0.205 & 0.233 & 0.324 & 0.452 &0.163  & 0.208 & 0.260 & \best{0.317}  \\
& & dom & \best{0.418} & \best{0.424} & 0.579 & \best{0.618} & \best{0.198} & 0.240 & \best{0.312} & \best{0.430} & \best{0.162} & \best{0.201} & \best{0.250} & \best{0.317} 

%% file: tables/ablate_shuf.tex
 & &\multicolumn{4}{c}{ETTh1} & \multicolumn{4}{c}{ETTm2} & \multicolumn{4}{c}{Weather} \\
& & 96 & 192 & 336 & 720 & 96 & 192 & 336 & 720 & 96 & 192 & 336 & 720  \\
\hline
\multirow{4}{*}{\rotatebox{90}{iTrans~\cite{liu2024itransformer}}}& Mask & 0.388 & 0.442 & 0.486 & 0.505
& 0.180 & 0.251 & \best{0.309} &  0.410 & 0.173 & 0.224 & 0.280 & 0.356   \\
& Noise & 0.387 & 0.445&0.482 &0.510 &0.180 &0.256 &0.312&0.409 &0.177	&0.222
	&0.281 & 0.359	\\
& Random & 0.386 & 0.440 & 0.479 & 0.499 & 0.183 & 0.254 & 0.311 & \best{0.407} & \best{0.171} & 0.222 & 0.280 & 0.358  \\
& Shuffle & \best{0.383} & \best{0.438} &  \best{0.473} & \best{0.492} & \best{0.178} & \best{0.246} & \best{0.309}  & 0.409 & \best{0.171} & \best{0.221} & \best{0.276} & \best{0.351}  \\
\hline
\multirow{4}{*}{\rotatebox{90}{MICN~\cite{wang2023micn}}}& Mask & 0.377 & \best{0.421} & 0.454 & 
0.543 & 0.175 & 0.268 & \best{0.337} & 0.505 & \best{0.178} & 0.239 & 0.283 & \best{0.342}    \\
& Noise &    0.393 & 0.430& 0.479 & 0.531 &0.201
   &0.331&0.366&0.561&0.201&0.236&0.281&0.351 \\
& Random & 0.381 & 0.423 & 0.476 & 0.670 & 0.183 & 0.284 & 0.367 & 0.614 & 0.182 & 0.233 & 0.282 & 0.349  \\
& Shuffle & \best{0.373} & \best{0.421} & \best{0.452} & \best{0.510} & \best{0.174} & \best{0.263} & 0.348 & \best{0.502} & 0.179 & \best{0.232} & \best{0.275} & \best{0.342} \\
\hline
\multirow{4}{*}{\rotatebox{90}{Lightts~\cite{zhang2022less}}}& Mask & 0.418 & 0.424 & 0.579 & 0.618 & 0.198 & 0.240 & \best{0.312} & 0.430 & \best{0.162} & \best{0.201} & 0.250& 0.317  \\
& Noise & 0.432&0.451  &0.566 
 &0.636&0.221&\best{0.236}&0.351&0.433&0.169&0.219&0.259&0.321	\\
& Random & 0.414 & 0.431 & 0.570 & 0.610 & 0.206 & 0.244 & 0.324& 0.442 & 0.171 & 0.213 & 0.263 & 0.323  \\
& Shuffle & \best{0.405} & \best{0.423} & \best{0.565} & \best{0.603} & \best{0.195} & 0.245 & \best{0.312} & \best{0.422} & 0.165 & 0.205& \best{0.249} & \best{0.312}

%% file: tables/long_term_6models_full_ett.tex
\multirow{11}{*}{\rotatebox{90}{iTransformer~\cite{liu2024itransformer}}} & Baseline & \base{0.392} & \base{0.447} & \base{0.483} & \base{0.516} & \base{0.303} & \base{0.381} & \base{0.412} & \base{0.434} & \base{0.344} & \base{0.383} & \base{0.421} & \base{0.494} & \base{0.183} & \base{0.251} & \base{0.311} & \base{0.412}  \\
& ASD~\cite{forestier2017generating} & 0.398 &0.456 &0.483 &0.512 &0.310 &0.388 &0.432 &0.452 &0.340 &0.382 &0.454&0.492&0.199	&0.254	&0.341&	0.423\\
& MSB~\cite{bandara2021improving} &0.387&0.460&0.494&0.531&0.309&0.382&0.447&0.433&0.339&0.386&0.467&0.510&0.187	&0.267	&0.332	&0.452\\
& Upsample~\cite{semenoglou2023data} & 0.391& 0.445&0.481&0.519&0.305&0.381&0.419&0.430&0.351&0.381&0.432&0.489&0.196	&0.279	&0.320 &	0.411\\
& FreqAdd~\cite{zhang2022self} & 0.389 & 0.446 & 0.475 & 0.510 & 0.300 & 0.384 & 0.416 & 0.438 & 0.350 & 0.385 & 0.422 & 0.490&0.187&	0.253&	0.311	&0.415
\\
& FreqPool~\cite{chen2023supervised} &0.433&0.456&0.497&0.532&0.313&0.392&0.415&0.450&0.347&0.392&0.430&0.499&0.187	&0.256	&0.324	&0.449
\\
& Robusttad~\cite{gao2020robusttad} &0.390&0.445&0.497&0.510&0.312&0.388&0.412&0.439&0.353&0.382&0.421&0.498&0.189	&0.255	&0.309	&0.428
\\
& STAug~\cite{zhang2023towards} & 0.390 & 0.445 &0.489 & 0.511& 0.323 & 0.428 & 0.486 & 0.483 & 0.339  & 0.383 & \best{0.417} & 0.485 &0.196	&0.267	&0.339	&0.449
\\
& MixMask~\cite{chen2023fraug} & 0.388 & 0.440 & 0.477 & 0.504 & 0.301 & \best{0.380} & 0.414 & 0.434 & 0.334 & 0.375 & 0.421 & \best{0.485} & \best{0.178} & 0.248 & 0.311 & \best{0.407}\\
& Ours& \best{0.383} & \best{0.438} & \best{0.473} & \best{0.492} & \best{0.298} & 0.382 & \best{0.411} & \best{0.428} & \best{0.332} & \best{0.374} & 0.424 & 0.492 & \best{0.178} & \best{0.246} & \best{0.309} & 0.409\\
\hline
\multirow{11}{*}{\rotatebox{90}{AutoFormer~\cite{wu2021autoformer}}} & Baseline& \base{0.429} & \base{0.440} & \base{0.495} & \base{0.498} & \base{0.381} & \base{0.443} & \base{0.471} & \base{0.475} & \base{0.467} & \base{0.610} & \base{0.529} & \base{0.773} & \base{0.233} & \base{0.278} & \base{0.383} & \base{0.488}  \\
& ASD & 0.450 & 0.485 & 0.523 & 0.556 & 0.370 & 0.465 & 0.476 & 0.503  & 0.480 & 0.620 & 0.502 & 0.633 &0.231&0.282&0.379&0.499\\
& MSB & 0.462&0.517& 0.612 & 0.579 &0.434 & 0.523 & 0.556 & 0.462 & 0.499 & 0.645 & 0.553 & 0.721 &0.232&0.285&0.389&0.487\\
& Upsample & 0.416 & 0.523 & 0.480 & \best{0.482} & 0.353 & 0.460 & 0.455 & 0.509 & 0.498 & 0.630 & 0.512& 0.667&0.234&0.291&0.382&0.521\\
& FreqAdd & 0.460 & 0.487 & 0.497 & 0.525 & 0.367 & 0.439 & 0.480 & 0.504 & 0.419 & 0.554 & 0.546 & 0.569 &0.223	&0.268&	0.330	&0.458
\\
& FreqPool &0.446&0.457&0.523&0.512&0.392&0.442&0.470&0.493&0.479&0.623&0.510&0.754&0.250&0.291&0.394&0.482\\
& Robusttad &0.437&0.452&0.492&0.477&0.367&0.497&0.502&0.527&0.432&0.510&0.553&0.623&0.235&	0.291&	0.375	&0.478
\\
& STAug & 0.429 & 0.478 & 0.505 & 0.506 & 0.354 & 0.443 & 0.496 & 0.495 & 0.415 & 0.581 & 0.588 & 0.693&0.224	&0.291	&0.338	&0.431
\\
& MixMask & 0.420 & 0.445 & 0.467 & 0.474 & 0.358 & 0.421 & 0.470 & 0.467 & 0.415 & 0.510 & 0.491 & 0.588 & 0.211 & 0.267 & 0.340 & 0.451 \\
& Ours& \best{0.409} & \best{0.436} & \best{0.458} & 0.486 & \best{0.335} & \best{0.419} & \best{0.453} & \best{0.452} & \best{0.392} & \best{0.506} & \best{0.491} & \best{0.559} & \best{0.210} & \best{0.266} & \best{0.329} & \best{0.429} \\
\hline
\multirow{11}{*}{\rotatebox{90}{MICN~\cite{wang2023micn}}} & Baseline& \base{0.384} & \base{0.425} & \base{0.464} & \base{0.574} & \base{0.358} & \base{0.518} & \base{0.566} & \base{0.827} & \base{0.313} & \base{0.360} & \base{0.389} & \base{0.461} & \base{0.200} & \base{0.282} & \base{0.375} & \base{0.606}  \\
& ASD &0.380&0.430&0.472&0.523&0.377&0.539&0.620&0.843&0.315&0.362&0.399&0.457&0.189	&0.331	&0.399&	0.617
\\
& MSB &0.423&0.423&0.501&0.559&0.402&0.623&0.790&1.126&0.330&0.358&0.402&0.459&0.192	&0.279 	&0.376	&0.651
\\
& Upsample &0.396&0.435&0.463&0.550&0.366&0.500&0.831&0.752&0.339&0.377&0.402&0.475&0.203	&0.291	&0.372&	0.595
\\
& FreqAdd & 0.390 & 0.430 & 0.477 & 0.643 & 0.370 & 0.521 & 0.626 & 0.975 & 0.316 & 0.360 & 0.407 & 0.478&0.176&	0.273	&0.378	&0.614
\\
& FreqPool &0.399&0.465&0.473&0.572&0.365&0.553&0.550&0.812&0.336&0.372&0.397&0.466&0.212&0.287&0.390&0.623\\
& Robusttad &0.392&0.436&0.491&0.556&0.339&0.529&0.553&0.998&0.339&0.359&0.396&0.472&0.200&0.296&0.356&0.617\\
& STAug & 0.374 & 0.429 & 0.489 & 0.608 & 0.413 & 0.760 & 1.330 & 2.608 & 0.313 & 0.360 & 0.418 & 0.483&0.180	&0.264	&0.323	&0.670
 \\
& MixMask & 0.378 & 0.423 & 0.461 & 0.521 & 0.339 & 0.488 & 0.544 & 0.735 & \best{0.301} & \best{0.352} & 0.401 & \best{0.454} & 0.183 & 0.278 & 0.356 & 0.528 \\
& Ours& \best{0.373} & \best{0.421} & \best{0.452} & \best{0.510} & \best{0.310} & \best{0.427} & \best{0.507} & \best{0.731} & 0.314 & 0.360 & \best{0.387} & 0.470 & \best{0.174} & \best{0.263} & \best{0.346} & \best{0.502} \\
\hline
\multirow{11}{*}{\rotatebox{90}{SCINet~\cite{liu2022scinet}}} & Baseline& \base{0.485} & \base{0.506} & \base{0.519} & \base{0.552} & \base{0.372} & \base{0.416} & \base{0.429} & \base{0.470} & \base{0.316} & \base{0.353} & \base{0.387} & \base{0.431} & \base{0.184} & \base{0.240} & \base{0.295} & \base{0.385}  \\
& ASD &0.494 & 0.480&0.491&0.559&0.362&0.402&0.432&0.499&0.331&0.367&0.389&0.453&0.197&	0.238	&0.296	&0.432
\\
& MSB &0.489 & 0.466&0.502&0.547&0.359&0.396&0.458&0.476&0.320&0.351&0.396&0.478&0.182	&0.237	&0.289	&0.449
\\
& Upsample & 0.471&0.457&0.479&0.541	&0.379&	0.407&	0.403&	0.482&	0.342&	0.386&	0.399	&0.442&0.179&	0.254&	0.292	&0.401
\\
& FreqAdd & 0.428 & 0.452 & 0.469 & 0.532 & 0.335 & 0.385 & 0.403 & 0.447 & 0.304 & 0.338 &0.373&0.421&0.174	&0.228	&0.286	&0.380 
\\
& FreqPool &0.499&0.510&0.557&0.549&0.410&0.453&0.432&0.475&0.331&0.362&0.379&0.432&0.185&0.239&0.302&0.399\\
& Robusttad &0.462&0.501&0.498&0.559&0.362&0.431&0.419&0.496&0.331&0.351&0.394&0.438&0.182&0.247&0.299&0.402\\
& STAug & 0.457 & 0.500 & 0.524 & 0.534 & 0.538 & 0.636 & 0.681 & 0.648 & 0.319 & 0.357 & 0.389 & 0.445&0.323	&0.407	&0.514	&0.668
\\
& MixMask & 0.427 & 0.452 & 0.465 & 0.548 & \best{0.335} & 0.377 & 0.400 & 0.438 & \best{0.302} & 0.341 & 0.376 & 0.423 & \best{0.174} & 0.230 & 0.289 & \best{0.368} \\
& Ours& \best{0.417} & \best{0.443} & \best{0.461} & \best{0.527} & \best{0.335} & \best{0.375} & \best{0.392} & \best{0.421} & \best{0.302} & \best{0.338} & \best{0.372} & \best{0.420}&\best{0.174}&\best{0.228}&\best{0.283}&0.372 \\
\hline
\multirow{11}{*}{\rotatebox{90}{TiDE~\cite{das2023longterm}}} & Baseline& \base{0.401} & \base{0.434} & \base{0.521} & \base{0.558} & \base{0.304} & \base{0.350} & \base{0.331} & \base{0.399} & \base{0.311} & \base{0.340} & \base{0.366} & \base{0.420} &0.166	&0.220	&0.273	&0.356
\\
& ASD &0.417&0.441&0.513&0.556&0.320&0.351&0.367&0.422&0.319&0.341&0.399&0.432&0.177&	0.241&	0.291&	0.371
\\
& MSB &0.422&0.476&0.529&0.579&0.331&0.379&0.334&0.401&0.302&0.356&0.382&0.451&0.182&	0.232&	0.287	&0.359
\\
& Upsample &0.431&0.452&0.533&0.604&0.346&0.372&0.350&0.456&0.324&0.339&0.378&0.463&0.203&	0.246	&0.306	&0.366
\\
& FreqAdd & 0.385 & 0.420 & 0.477 & 0.505 & 0.289 & 0.336 & 0.330 & 0.390 & 0.309 & 0.339 &0.365&0.417&0.164&	0.219	&0.273	&0.355
\\
& FreqPool &0.423&0.455&0.510&0.592&0.312&.376&0.339&0.397&0.319&0.352&0.397&0.453&0.179&0.231&0.299&0.371\\
& Robusttad &0.396&0.432&0.521&0.537&0.331&0.352&0.337&0.398&0.321&0.346&0.382&0.437&0.180&0.225&0.282&0.371\\
& STAug & 0.515 & 0.535 & 0.521 & 0.558 & 0.390 & 0.437 & 0.403 & 0.508 & 0.310 & 0.337 & \best{0.364} & 0.417&0.222	&0.343	&0.515&	0.847
\\
& MixMask & \best{0.385} & 0.420 & 0.478 & 0.507 & 0.289 & 0.339 & 0.330 & 0.391 & 0.299 & 0.332 & 0.367 & 0.416 & \best{0.165} & 0.219 & \best{0.271} & \best{0.347} \\
& Ours& \best{0.385} & \best{0.414} & \best{0.467} & \best{0.498} & \best{0.283} & \best{0.332} & \best{0.324} & \best{0.388} & \best{0.297} & \best{0.328} & \best{0.365} & \best{0.412} & \best{0.165} & \best{0.218} & \best{0.271} & 0.350 \\
\hline
\multirow{11}{*}{\rotatebox{90}{LightTS~\cite{zhang2022less}}} & Baseline& \base{0.448} & \base{0.444} & \base{0.663} & \base{0.706} & \base{0.369} & \base{0.476} & \base{0.738} & \base{1.165} & \base{0.323} & \base{0.347} & \base{0.428} & \base{0.476} & \base{0.212} & \base{0.237} & \base{0.350} & \base{0.473}  \\
& ASD &0.451&0.476&0.633&0.681&0.392&0.469&0.701&0.998&0.356&0.352&0.441&0.478&0.258	&0.251&	0.351	&0.483
\\
& MSB &0.467&0.463&0.627&0.652&0.378&0.472&0.652&1.123&0.371&0.349&0.430&0.479&0.236	&0.242	&0.359	&0.471
\\
& Upsample &0.449&0.472&0.610&0.637&0.401&0.487&0.714&1.245&0.329&0.366&0.453&0.492&0.241	&0.255	&0.366	&0.492
\\
& FreqAdd & 0.417 & 0.430 & 0.578 & 0.622 & 0.351 & 0.453 & 0.689 & 1.125 & 0.322 &0.352&0.400&0.450&0.206&	0.237	&0.327	&0.455
\\
& FreqPool &0.463&0.471&0.652&0.690&0.369&0.512&0.723&1.264&0.336&0.351&0.442&0.497&0.233&0.259&0.372&0.453\\
& Robusttad &0.445&0.442&0.590&0.654&0.372&0.468&0.699&0.982&0.331&0.352&0.441&0.462&0.232	&0.227	&0.342	&0.446
\\
& STAug & 0.445 & 0.441 & 0.669 & 0.714 & 0.520 & 0.807 & 2.101 & 2.467 & 0.320 & 0.343 & 0.427 & 0.476&0.230 	&0.266	&0.372&	0.475
\\
& MixMask & 0.417 & 0.429 & 0.575 & 0.620 & 0.337 & 0.426 & 0.643 & 0.993 & \best{0.316} & \best{0.340} & 0.398 & 0.447 & 0.199 & \best{0.233} & 0.322 & 0.440 \\
& Ours& \best{0.405} & \best{0.423} & \best{0.565} & \best{0.603} & \best{0.335} & \best{0.395} & \best{0.575} & \best{0.827} & 0.322 & \best{0.340} & \best{0.391} & \best{0.440} & \best{0.195} & 0.245 & \best{0.312} & \best{0.422}

%% file: tables/long_term_6models_full_other4.tex
\multirow{11}{*}{\rotatebox{90}{iTransformer~\cite{liu2024itransformer}}} & Baseline& \base{0.152} & \base{0.159} & \base{0.179} & \base{0.230} & \base{0.175} & \base{0.224} & \base{0.281} & \base{0.362} & \base{\best{0.086}} & \base{0.180} & \base{0.335} & \base{0.856} & \base{0.399}&	\base{0.418}	&\base{0.428}&	\base{0.463}
\\
& ASD~\cite{forestier2017generating} &0.173&	0.179&	0.201	&0.234&0.191 &	0.223	&0.280 &	0.364&	0.088&	0.183&	0.343&	0.872&0.431&0.428&0.430&0.478\\
& MSB~\cite{bandara2021improving} &0.182	&0.182 &	0.194&	0.267&0.185	&0.235&	0.284&	0.359 &	0.089	&0.189	&0.359	&0.907&0.417	&0.416	&0.422	&0.471
\\
& Upsample~\cite{semenoglou2023data} &0.166	&0.188&	0.216	&0.221&0.204	&0.257&	0.291	&0.373&	0.086&	0.180 	&0.338 	&0.834&0.433&0.419&0.433&0.476\\
& FreqAdd~\cite{zhang2022self} & 0.150 & 0.157 & 0.172 & 0.204 & 0.181 & 0.230 & 0.285 & 0.362 & 0.087 & 0.181 
    &0.333 &  0.837  & 0.480	&0.441	&0.450	&0.501
\\
& FreqPool~\cite{chen2023supervised} &0.169&0.170&0.194&0.237&0.184&0.223&0.279&0.378&0.088&0.183&0.330&0.832&0.410 	&0.429	&0.433	&0.476
\\
& Robusttad~\cite{gao2020robusttad} &0.150&0.157&0.176&0.210&0.172&0.225&0.281&0.357&0.087&0.179&0.329&0.833&0.406&	0.417	&0.429	&0.458
\\
& STAug~\cite{zhang2023towards} & 0.160 & 0.173 & 0.218 & 0.372 & 0.206 & 0.264 & 0.319 & 0.385 & 0.086 & 0.178 &0.335 & 0.866&0.413	&0.432	&0.449	&0.481
\\
& MixMask~\cite{chen2023fraug} & 0.151 & 0.158 & 0.173 & 0.205 & 0.175 & 0.224 & 0.279 & 0.354 & 0.089 & 0.178 & 0.328 & 0.845&0.395	&\best{0.401}&	\best{0.418}	&0.450
\\
& Ours& \best{0.150} & \best{0.156} & \best{0.171} & \best{0.199} &  \best{0.171} & \best{0.221} & \best{0.276} & \best{0.351} & \best{0.086} & \best{0.176} & \best{0.313} & \best{0.821}&\best{0.394}&	0.412	&0.423&	\best{0.448}
\\
\hline
\multirow{11}{*}{\rotatebox{90}{AutoFormer~\cite{wu2021autoformer}}} & Baseline& \base{0.203} & \base{0.208} & \base{0.231} & \base{0.239} & \base{0.241} & \base{0.314} & \base{0.341} & \base{0.425} & \base{0.143} & \base{0.305} & \base{0.470} & \base{1.056}  & \base{0.640} & \base{0.645} & \base{0.611} & \base{0.658}\\
& ASD &0.247&	0.216	&0.221	&0.235&0.652&	0.392&	0.416	&0.513&	0.141&	0.280 &	0.579	&1.240&0.631&	0.602	&0.607	&0.643
\\
& MSB &0.237	&0.256&	0.295	&0.236&0.256	&0.379	&0.402&	0.468&	0.156&	0.254	&0.513	&1.339&0.652	&0.665&	0.643	&0.65
\\
& Upsample &0.201&	0.209	&0.232	&0.268&0.281&	0.294&	0.329	&0.385	&0.141	&0.292	&0.553&	1.295&0.653	&0.676	&0.702	&0.694
\\
& FreqAdd & 0.193 & 0.197 & 0.212 & 0.225 & 0.255 & 0.323 & 0.370 & 0.419 & 0.143 & 0.369 & 0.716 & 1.173 &0.613	&0.598 &	0.617 	&0.639
\\
& FreqPool &0.213&	0.224&	0.234&	0.257&0.237&	0.339	&0.372&	0.446&	0.142&	0.336	&0.532&	1.014&0.63	&0.598	&0.603	&0.639
\\
& Robusttad &0.230 &	0.242	&0.261	&0.231&0.27&	0.334&	0.351&	0.429	&0.142	&0.309	&0.462	&1.123&0.621	&0.614	&0.612	&0.646
\\
& STAug & 0.191 & 0.206 & 0.217 & 0.234 & 0.250 & 0.300 & 0.347 & 0.418 & 0.140 & 0.326 & 0.594 & 1.176&0.632	&0.619	&0.632	&0.640
\\
& MixMask & 0.177 & 0.194 & 0.206 & 0.224 & 0.240 & 0.302 & 0.330 & 0.422 & 0.141 & 0.284 & 0.453 & 0.778& \best{0.560}	&0.584	&0.594	&\best{0.635}
\\
& Ours& \best{0.171} & \best{0.191} & \best{0.203} & \best{0.219}  & \best{0.214} & \best{0.273} & \best{0.327} & \best{0.383} & \best{0.136} & \best{0.243} & \best{0.418} & \best{0.695}& 0.577&	\best{0.581}&\best{	0.592}	&0.638
\\
\hline
\multirow{11}{*}{\rotatebox{90}{MICN~\cite{wang2023micn}}} & Baseline& \base{0.171} & \base{0.183} & \base{0.198} & \base{0.224} & \base{0.188} & \base{0.241} & \base{0.278} & \base{0.350} & \base{0.091} & \base{0.185} & \base{0.355} & \base{0.941} & \base{0.522} & \base{0.540} & \base{0.553} & \base{0.573} \\
& ASD &0.165	&0.174&	0.190	&0.237&0.189&	0.242	&0.276&	0.354	&0.087	&0.175 	&0.337	&1.203&0.505&	0.534	&0.541	&0.539
\\
& MSB &0.179	&0.182&	0.201	&0.225&0.201&	0.250 &	0.291&	0.365&	0.088	&0.176&	0.360 &	0.995&0.513	&0.532&	0.528	&0.556
\\
& Upsample &0.182	&0.180 &	0.203	&0.220 &0.193	&0.249	&0.279&	0.372	&0.084&	0.171	&0.313	&\best{0.702}&0.533&	0.559&	0.556	&0.590
\\
& FreqAdd & 0.160 & 0.169 & 0.182 & 0.199 & 0.180 & 0.234 & 0.282 & 0.350 & 0.087 & 0.174 & 0.349 & 0.923 &0.503&0.527&0.520&0.571\\
& FreqPool &0.182&	0.203&	0.241	&0.256&0.192	&0.257	&0.278	&0.351	&0.089	&0.179	&0.394	&0.923&0.531&0.539&0.556&0.592\\
& Robusttad &0.179&	0.220& 	0.234&	0.227&0.192	&0.239&	0.292&	0.343&	0.085&	0.179	&0.336&	0.932&0.510	&0.532	&0.547	&0.597
\\
& STAug & 0.180 & 0.195 & 0.210 & 0.224 & 0.272 & 0.356 & 0.433 & 0.559 & 0.092 & 0.183 & 0.313 & 0.790 &0.512&0.533&0.529&0.585\\
& MixMask & 0.159 & \best{0.165} & \best{0.178} & \best{0.195} & 0.185 & 0.239 & 0.281 & 0.344 & 0.086 & 0.174 & 0.337 & 0.796&\best{0.490}	&0.512	&0.519&	\best{0.538}
 \\
& Ours& \best{0.157} & 0.168 & \best{0.178} & 0.211 & \best{0.179} & \best{0.232} & \best{0.275} & \best{0.342} & \best{0.084} & \best{0.169} & \best{0.303} & 0.750&0.501	&\best{0.507}	&\best{0.518}	&0.556
 \\
\hline
\multirow{11}{*}{\rotatebox{90}{SCINet~\cite{liu2022scinet}}} & Baseline& \base{0.212} & \base{0.237} & \base{0.255} & \base{0.286} & \base{0.229} & \base{0.282} & \base{0.334} & \base{0.402} & \base{0.099} & \base{0.191} & \base{0.356} & \base{0.916}  & \base{0.550} & \base{0.526} & \base{0.545} & \base{0.596}\\
& ASD &0.229&	0.241&	0.239	&0.282&0.254	&0.276&	0.356&	0.462&	0.095	&0.204&	0.379	&1.230&0.537&	0.521	&0.541&	0.570 
\\
& MSB &0.232&	0.237&	0.228	&0.274&0.279	&0.265	&0.374	&0.454	&0.093	&0.267&	0.402	&0.965&0.520&	0.510 &	0.537	&0.565
\\
& Upsample &0.250	&0.232	&0.271&	0.309&0.243	&0.299&	0.361	&0.431	&0.092&	0.196&	\best{0.311}	&0.932&0.519	&0.536	&0.528	&0.576
\\
& FreqAdd & 0.176 & 0.195 & 0.212 & 0.237 & 0.208 & 0.258 & 0.309 & 0.385 & 0.092 & 0.186 & 0.343 & 0.920&\best{0.492}	&0.497	&0.512&	0.550 
\\
& FreqPool &0.230	&0.221	&0.242	&0.339&0.261	&0.290 &	0.337&	0.456	&0.096	&0.183	&0.551&	0.938&0.557&0.519&0.533&0.562\\
& Robusttad &0.189&	0.202	&0.210 	&0.243&0.229&	0.281	&0.331&	0.410 &	0.093	&0.186	&0.334	&0.957&0.523&0.519&0.522&0.569\\
& STAug & 0.210&0.239 & 0.282 & 0.411 & 0.277 & 0.329  & 0.372 & 0.435 & 0.098 & 0.191 & 0.342 & 0.931 &0.560&	0.517 &	0.521	&0.566
\\
& MixMask & \best{0.171} & \best{0.188} & 0.204 & 0.230  & 0.205 & 0.250 & 0.310 & \best{0.374} & 0.093 & 0.179 & 0.336 & 0.928&0.495&\best{0.492}	&0.511	&0.551
 \\
& Ours& 0.172 & \best{0.188} & \best{0.200} & \best{0.225} & \best{0.197} & \best{0.246} & \best{0.299} & 0.379 & \best{0.091} & \best{0.175} & 0.342 & \best{0.890}& 0.500	&0.495&	\best{0.509}	&\best{0.544}
\\
\hline
\multirow{11}{*}{\rotatebox{90}{TiDE~\cite{das2023longterm}}} & Baseline& \base{0.207} & \base{0.197} & \base{0.211} & \base{0.238} & \base{0.177} & \base{0.220} & \base{0.265} & \base{0.323} & \base{0.093} & \base{0.184} & \base{0.330} & \base{0.860} & \base{0.452} & \base{0.450} & \base{0.451} & \base{0.479} \\
& ASD &0.232&0.220&0.231&0.265&0.189&0.221&0.297&0.332&0.095&0.206&0.351&0.962&0.477&0.462&0.450&0.506\\
& MSB &0.210&0.219&0.253&0.261&0.199&0.254&0.273&0.339&0.092&0.179&0.358&0.941&0.461&0.451&0.455&0.510\\
& Upsample &0.206&0.199&0.223&0.274&0.203&0.267&0.331&0.355&0.091&0.182&0.331&0.852&0.490&0.466&0.472&0.493\\
& FreqAdd & 0.150 & 0.163 & 0.177 & 0.209 & \best{0.173} & \best{0.216} & 0.263 & \best{0.322} & \best{0.088} & 0.180 & 0.330 & 0.848 &0.429&0.441&0.440&0.471\\
& FreqPool &0.224&0.238&0.233&0.270&0.189&0.224&0.292&0.334&0.092&0.334&0.521&1.124&0.453&0.466&0.479&0.503\\
& Robusttad &0.176&0.166&0.182&0.229&0.182&0.231&0.279&0.330&0.099&0.232&0.331&0.924&0.449&0.430&0.438&0.482\\
& STAug & 0.230 & 0.210& 0.192 & 0.225 & 0.205 & 0.247 & 0.292 & 0.364 & 0.092 & 0.184 & 0.330 & 0.859 &0.466&0.455&0.471&0.480\\
& MixMask & \best{0.143} & 0.155 & \best{0.164} & 0.210 & \best{0.173} & \best{0.216} & 0.263 & 0.323 & 0.089 & 0.180 & 0.329 & 0.861 &\best{0.421}&0.427&0.434&\best{0.466}\\
& Ours& \best{0.143} & \best{0.150} & 0.165 & \best{0.202} & 0.177 & 0.219 & \best{0.261} & \best{0.322} & \best{0.088} & \best{0.179} & \best{0.324} & \best{0.847}&0.423&\best{0.426}&\best{0.433}&\best{0.466} \\
\hline
\multirow{11}{*}{\rotatebox{90}{LightTS~\cite{zhang2022less}}} & Baseline& \base{0.210} & \base{0.169} & \base{0.182} & \base{0.212} & \base{0.168} & \base{0.210} & \base{0.260} & \base{0.320} & \base{0.139} & \base{0.252} & \base{0.412} & \base{0.840}  & \base{0.505} & \base{0.515} & \base{0.539} & \base{0.587}\\
& ASD &0.225	&0.179	&0.198&	0.232&0.179	&0.21	&0.271&	0.321&	0.132	&0.320 &	0.436&	1.036&0.510 	&0.514	&0.534	&0.579
\\
& MSB &0.233	&0.182	&0.204	&0.228&0.170	&0.214&	0.259	&0.332&	0.117&	0.294	&0.502	&0.964&0.532	&0.510 	&0.539	&0.584
\\
& Upsample &0.246 &	0.179	 &0.211 &	0.254 &0.182	 &0.223	 &0.257	 &0.336	 &0.099 &	0.251 &	0.369	 &0.702&0.522	&0.547&	0.532	&0.597
\\
& FreqAdd & 0.213 & 0.159 & 0.177 & 0.210 & 0.164 & 0.207 & 0.258 & 0.317 & 0.098 & 0.522 & 0.565 & 1.583&0.492	&0.500 	&0.530 &	0.572
\\
& FreqPool &0.219&	0.174	&0.197&	0.236&0.193	&0.254	&0.267&	0.339&	0.099	&0.275&	0.394&	0.793&0.501&0.519&0.533&0.592\\
& Robusttad &0.212&	0.169&	0.181&	0.223&0.172&	0.223	&0.259	&0.324	&0.092	&0.279	&0.451	&0.796&0.499&0.502&0.521&0.572\\
& STAug & 0.224 & 0.267 & 0.294 & 0.351 & 0.214 & 0.263 & 0.382 & 0.371 & 0.096 & \best{0.212} & 
0.380 & 0.690&0.520 &	0.534&	0.520 &	0.596
 \\
& MixMask & \best{0.192} & 0.158 & 0.175 & 0.211 & \best{0.163} & 0.206 & 0.257 & 0.318 & 0.099 & 0.384 & 0.518 & 0.774&0.486	&0.499&	0.517	&\best{0.555}
 \\
& Ours& 0.210 & \best{0.156} & \best{0.173} & \best{0.206} & 0.165 & \best{0.205} & \best{0.249} & \best{0.312} & \best{0.088} & 0.243 & \best{0.361} & \best{0.676}&\best{0.483}&	\best{0.497}&	\best{0.515}	&0.567

%% file: tables/short_term_1model_full.tex
\multirow{2}{*}{Methods} & \multicolumn{4}{c}{PEMS03} & \multicolumn{4}{c}{PEMS04} & \multicolumn{4}{c}{PEMS07} & \multicolumn{4}{c}{PEMS08}\\
& 12 & 24 & 36 & 48 & 12 & 24 & 36 & 48 & 12 & 24 & 36 & 48& 12 & 24 & 36 & 48 \\
\hline
 Baseline& \base{0.070} & \base{0.097} & \base{0.134} & \base{0.164} & \base{0.088} & \base{0.124} & \base{0.160} & \base{0.196} & \base{0.067} & \base{0.097} & \base{0.128} & \base{0.156}  & \base{0.088} & \base{0.136} & \base{0.191} & \base{0.248}\\
ASD~\cite{forestier2017generating} &0.072&0.096&0.152&0.239&0.098&0.132&0.156&0.190&0.069&0.099&0.154&0.181&0.089&0.138&0.196&0.247\\
MSB~\cite{bandara2021improving} &0.096&0.131&0.129&0.214&0.087&0.134&0.167&0.219&0.098&0.096&0.137&0.165&0.096&0.137&0.210&0.256\\
Upsample~\cite{semenoglou2023data} &0.069&0.096&0.128&0.179&0.087&0.124&0.158&0.199&0.072&0.099&0.127&0.155&0.088&0.140&0.192&0.245\\
FreqAdd~\cite{zhang2022self} &1.036	&0.104	&0.251&	0.362&0.088	&0.125	&0.159&	0.201	&0.067&	0.097	&0.127&	0.155&0.089	&0.135	&0.192	&0.253
\\
FreqPool~\cite{chen2023supervised} &1.234	&0.178	&0.296&	0.451	&0.099	&0.145	&0.178	&0.226	&0.079	&0.104	&0.152	&0.172&0.099&0.155&0.203&0.264\\
Robusttad~\cite{gao2020robusttad} &0.082&	0.098&	0.132&	1.520&	0.089&	0.123	&0.161	&0.195	&0.067&	0.097	&0.129	&0.157&0.092&	0.135	&0.189&	0.26
\\
STAug~\cite{zhang2023towards} &0.079&	0.112&	0.195&	0.456	&0.087&	0.120 &	0.162	&0.304	&0.066	&0.096	&0.132	&0.165&0.092&	0.147	&0.192	&0.276
\\
Mask~\cite{chen2023fraug} & 0.443 & 1.205 & 0.233 & 1.510 & 0.086 & 0.119 & 0.158 & 0.346 & \best{0.065} & 0.095 & 0.125 & 0.156&0.089&	\best{0.131}&	0.186	&0.239
 \\
Mix~\cite{chen2023fraug} & 1.018 & 0.097 & 0.877 & 1.501 & \best{0.085} & 0.119 & 0.154 & 0.205 & \best{0.065} & \best{0.094} & 0.134 & 0.152&0.089&	\best{0.131}&	\best{0.184}	&\best{0.234}
\\
Ours& \best{0.067} & \best{0.095} & \best{0.126} & 0.235 &  \best{0.085} & \best{0.118} & \best{0.149} & \best{0.182} & \best{0.065} & \best{0.094} & \best{0.123} & \best{0.148}&\best{0.087}	&0.134&	\best{0.184}	&0.240